\documentclass[acmtog]{acmart}

\usepackage{booktabs} % For formal tables

\usepackage{amsmath,amsfonts}

\usepackage{array}
\usepackage{caption}
\usepackage{graphicx}%, subfig}
\usepackage{enumerate}
\usepackage{subfigure}
\graphicspath{{figure/}{./}} 
\usepackage{multirow}
\usepackage{amsmath}
\usepackage{textcomp}
\usepackage{stfloats}
\usepackage{url}
\usepackage{verbatim}
\usepackage[ruled,linesnumbered]{algorithm2e}

\SetAlFnt{\small}
\SetAlCapFnt{\small}
\SetAlCapNameFnt{\small}
\SetAlCapHSkip{0pt}

\begin{document}
% Title portion
\title{Guided Linear Upsampling}
%for Real-time Interactive Image Editing}

% DO NOT ENTER AUTHOR INFORMATION FOR ANONYMOUS TECHNICAL PAPER SUBMISSIONS TO SIGGRAPH 2019!

\author{Shuangbing Song}
\orcid{0000-0002-1031-6151}
% \affiliation{%
%  \institution{Shandong University}
%  \streetaddress{72 Binhai Rd}
%  \city{Qingdao}
%  \state{Shandong}
%  \postcode{266200}
\email{songs@mail.sdu.edu.cn}
\author{Fan Zhong}
\orcid{0000-0001-7636-524X}
\authornote{Corresponding author.}
% \affiliation{%
%  \institution{Shandong University}
%  \streetaddress{72 Binhai Rd}
%  \city{Qingdao}
%  \state{Shandong}
%  \postcode{266200}
\email{zhongfan@sdu.edu.cn}
\author{Tianju Wang}
\orcid{0009-0009-2028-5051}
% \affiliation{%
%  \institution{Shandong University}
%  \streetaddress{72 Binhai Rd}
%  \city{Qingdao}
%  \state{Shandong}
%  \postcode{266200}
\email{202000130103@mail.sdu.edu.cn}
\author{Xueying Qin}
\orcid{0000-0003-0057-295X}
% \affiliation{%
%  \institution{Shandong University}
%  \streetaddress{1500 Shunhua Rd}
%  \city{Jinan}
%  \state{Shandong}
%  \postcode{250101}
\email{qxy@sdu.edu.cn}
\author{Changhe Tu}
\orcid{0000-0002-1231-3392}
\email{chtu@sdu.edu.cn}
\affiliation{%
 \institution{Shandong University}
 % \streetaddress{72 Binhai Rd}
 % \city{Qingdao}
 % \state{Shandong}
 % \postcode{266200}
 \country{China}
 }

%\affiliation{%
%  \institution{Inria Paris-Rocquencourt}
%\email{beranger@inria.fr}
%\affiliation{%
% \institution{Rajiv Gandhi University}
%\email{aprna_patel@rguhs.ac.in}
%\affiliation{%
%  \institution{Tsinghua University}
%  \streetaddress{30 Shuangqing Rd}
%  \country{China}
%}
%\email{chan0345@tsinghua.edu.cn}
%\author{Ting Yan}
%\affiliation{%
%  \institution{Eaton Innovation Center}
%  \city{Prague}
%  \country{Czech Republic}}
%\email{yanting02@gmail.com}
%\author{Tian He}
%\affiliation{%
%  \institution{University of Virginia}
%  \department{School of Engineering}
%  \city{Charlottesville}
%  \state{VA}
%  \postcode{22903}
%  \country{USA}
%}
%\affiliation{%
%  \institution{University of Minnesota}
%\email{tinghe@uva.edu}
%\author{Chengdu Huang}
%\author{John A. Stankovic}
%\author{Tarek F. Abdelzaher}
%\affiliation{%
%  \institution{University of Virginia}
%  \department{School of Engineering}
%  \city{Charlottesville}
%  \state{VA}
%  \postcode{22903}
%  \country{USA}
%}

%\renewcommand\shortauthors{Zhou, G. et al}

\begin{abstract}

Guided upsampling is an effective approach for accelerating high-resolution image processing. In this paper, we propose a simple yet effective guided upsampling method. Each pixel in the high-resolution image is represented as a linear interpolation of two low-resolution pixels, whose indices and weights are optimized to minimize the upsampling error. The downsampling can be jointly optimized in order to prevent missing small isolated regions. Our method can be derived from the \emph{color line model} and \emph{local color transformations}. Compared to previous methods, our method can better preserve detail effects while suppressing artifacts such as bleeding and blurring. It is efficient, easy to implement,  and free of sensitive parameters. We evaluate the proposed method with a wide range of image operators, and show its advantages through quantitative and qualitative analysis. We demonstrate the advantages of our method for both interactive image editing and real-time high-resolution video processing. In particular, for interactive editing, the joint optimization can be precomputed, thus allowing for instant feedback without hardware acceleration.

\end{abstract}

\setcopyright{acmlicensed}
\acmJournal{TOG}
\acmYear{2023} \acmVolume{42} \acmNumber{4} \acmArticle{} \acmMonth{8} \acmPrice{15.00}\acmDOI{10.1145/3592453}
% \if0
%\begin{comment}


% \fi
%\end{comment}
% \keywords{High-resolution, image editing, downsampling, upsampling}

\newcommand{\downscale}{{_\downarrow}}
\newcommand{\Ilow}{I^\downscale}
\newcommand{\Tlow}{T^\downscale}
\newcommand{\refeq}[1]{Eq. (\ref{#1})}
\newcommand{\reffig}[1]{Figure \ref{#1}}
\newcommand{\Omegaplow}{\Omega_{p^\downscale}}

\newcommand{\modified}[1]{#1}
\newcommand{\modifiedbeg}{}
\newcommand{\modifiedend}{}

\newcommand{\revised}[1]{#1}
\newcommand{\revisedbeg}{}
\newcommand{\revisedend}{}

\newcommand{\onerevise}[1]{#1}
\newcommand{\onerevisebeg}{}
\newcommand{\onereviseend}{}

\begin{teaserfigure}
	\centering 
	\includegraphics[width=0.98\textwidth]{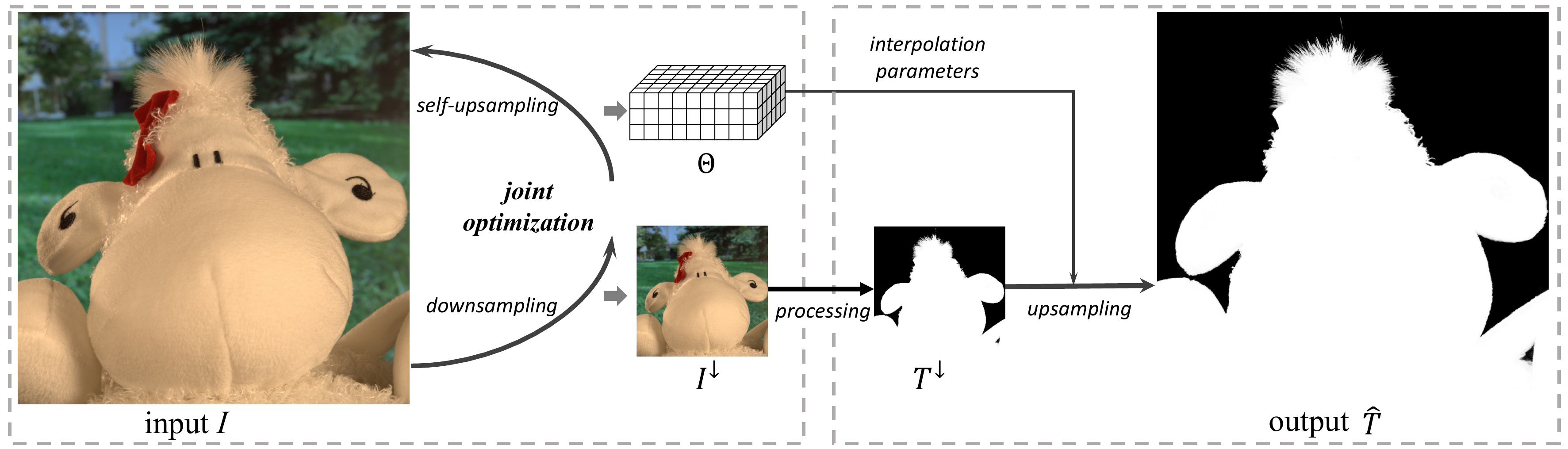}
	\caption{\revised{Our method to accelerate high-resolution image processing with \onerevise{guided} linear upsampling. Given a high-resolution source image $I$, our method can jointly optimize the downsampled source image $\Ilow$ and the interpolation parameters $\Theta$, and then $\Ilow$ is processed by a black-box image operator to get the low-resolution target image $\Tlow$. The high-resolution target image $\hat{T}$ can be linearly upsampled from $\Tlow$ with the optimized parameters $\Theta$.}}
	\label{fig:intro}
\end{teaserfigure}

%	\centering 
%	\subfigure[8$\times\downarrow$ input with guidance]{
%		\includegraphics[width=0.205\linewidth]{figure/fig2/2-fig0816.pdf}
%	} 
%	\subfigure[Ground truth]{
%		\includegraphics[width=0.18\linewidth]{figure/fig2/target0816.png}
%	} 
%	\subfigure[JBU($\sigma_d=0.5,\sigma_r=0.1$)]{ 
%		\includegraphics[width=0.18\linewidth]{figure/fig2/jbu0501.png}
%	}
%	\subfigure[BGU]{
%		\includegraphics[width=0.18\linewidth]{figure/fig2/bgu0816.png} 
%	} 
%	\subfigure[GLU]{
%		\includegraphics[width=0.18\linewidth]{figure/fig2/glu0816.png} 
%	} 
%
%	\caption{Comparison of the representative guided upsampling operators. JBU~\cite{kopf2007joint} tends to over blur the target image, BGU~\cite{chen2016bilateral} is prone to lose detail and produce bleeding effects. Our method can represent the pixel-level editing effects of the target image.} 
%	\label{fig:downsample} 

\maketitle

\section{Introduction}
\label{sec:intro}

In the past decades, many useful image processing methods have been proposed for various tasks such as enhancement~\cite{aubry2014fast}, style transfer~\cite{CycleGAN2017,li2018closed}, matting~\cite{levin2007closed}, colorization~\cite{iizuka2016let}, etc. Most of them require intensive computation and memory, and thus face great challenges for high-resolution images. At the same time, 
the popularity of mobile devices requires us to consider more about computational efficiency. The problem is even more prominent for interactive image editing~\cite{levin2004colorization,bousseau2009user}, which requires repetitive user interactions, so instant feedback is necessary for a better user experience. 

For general image processing, the \emph{guided upsampling} should be the simplest and most effective way to achieve acceleration. By using the original image as a guidance map, a large ratio downsampling of the output image can be upsampled to the original resolution without noticeable artifacts. This is amazing because even for image operators of linear complexity in image size, using $8\times$ downsampling can result in $64\times$ speed up. %without improving the algorithm. 

Two classical approaches for guided upsampling are \emph{joint bilateral upsampling} (JBU)~\cite{kopf2007joint} and \emph{bilateral guided upsampling} (BGU)~\cite{chen2016bilateral}. JBU is an extension of the bilateral filter~\cite{paris2006fast, durand2002fast}, while BGU is based on the local color transformations~\cite{levin2007closed}, whose effectiveness for guided upsampling has been demonstrated in earlier works such as \emph{transform \onerevise{recipes}}~\cite{gharbi2015transform} and \emph{guided filter}~\cite{he2012guided}. In BGU, the local transformations are applied in the bilateral space~\cite{barron2015fast}, which further improves the efficiency and quality. Recent works are mainly learning-based~\cite{gharbi2017deep,xia2020joint,xia2021real,dai2021learning}, which can better employ domain knowledge for improving quality. However, they need to be trained for each specific task, and thus cannot be generalized to other tasks. Instead, we will follow the roadmap of the classical approaches, in order to seek a universal guided upsampler applicable to a wide range of image operators.

In this paper, we propose \emph{Guided Linear Upsampling} (GLU), which is pretty simple but very effective. We introduce a new representation of high-resolution images, with each pixel represented as the linear interpolation of only two low-resolution pixels. By optimizing the representation parameters, i.e. the indices and weights of the interpolated pixel pairs, very small errors can be achieved even for large ratio upsampling. The parameters can be optimized for the source image and then applied to the target image, resulting in a high-resolution target image that can well preserve details while avoiding artifacts such as bleeding and blurring. We also propose an efficient method to optimize the downscaled source image, in order to better preserve thin structures and small regions. As illustrated in \reffig{fig:intro}, with our method the downsampling and upsampling can be jointly optimized in order to minimize the upsampling error, thus effectively preventing the loss of small isolated local structures.

\revised{The proposed method contains \onerevise{only a few} parameters that are easy to set, and a fixed parameter setting can be well suited for various tasks and input images. It is efficient and easy to \onerevise{implement}, and can achieve fast speed with a GPU implementation. Moreover, the joint optimization of upsampling and downsampling is \emph{target-free}, i.e. independent of the target image, and thus needs to be done only once for each image, regardless of how the target image changes. As a demonstration, we show that with our method, real-time image editing and video processing can be achieved  easily for time-costly image operators. }

\section{Related Work}
\label{sec:relatedwork}

The computational efficiency of image processing algorithms is very important for many applications, so it has attracted a lot of studies for various image processing tasks, such as filtering~\cite{vaudrey2009fast,hengsheng2011tone}, enhancement~\cite{farbman2008edge} and modern learning-based image synthesis~\cite{chai2022any} etc. GPU-based acceleration has also been widely studied~\cite{wu2009improved, kazhdan2008streaming, li2012hdr}. Most of these methods are specific to \onerevise{the} processing algorithms. In contrast, we will focus on guided upsampling, which can accelerate a wide range of image operators by treating them as black boxes.

The problem of guided upsampling is first introduced in \cite{kopf2007joint}, in which \emph{joint bilateral upsampling} (JBU) is proposed as the solution. JBU represents each output pixel as the weighted average of a set of low-resolution pixels. The weights are computed with bilateral weighting function~\cite{tomasi1998bilateral} incorporating the guidance of the high-resolution input image. As a result, JBU also inherits the problems of the bilateral filter, such as edge blur and gradient reversal~\cite{durand2002fast,he2012guided}. Our method has the same general form as JBU, but it involves the weighted average of only two pixels, and the artifacts can be avoided by the proposed optimization techniques.

In \emph{guided image filtering}~\cite{he2012guided}, the target image is locally represented as the affine transformations of the source image. This approach is effective in preserving the local structures of the source image. \cite{gharbi2015transform} introduces the concept of \emph{transform recipe} for efficient cloud image processing, which shows that high-quality upsampling can be obtained with local affine transformations for a wide range of processing tasks. 
\cite{chen2016bilateral} proposes to apply the local color transformations in the bilateral space, which further enhances the ability to represent detail effects by localizing the transformations in both spatial and range spaces.

Recent works mainly achieve improvements by leveraging machine learning. \cite{gharbi2017deep} shows that the local color transformations can be directly learned from pairs of training images, which eliminates the need to run the processing operator online. \cite{wu2018fast} proposes an end-to-end trainable guided filter by formulating the local transformations into a fully differentiable module. For better preserving details, \cite{pan2019spatially} proposes to use  linear representations with more localized support and learning-based regularization. \cite{shi2021unsharp} reformulates the guided filter as an unsharp mask operator more suitable for learning. Although impressive results can be achieved, the learning-based methods need to be trained for each specific task, and thus may be unavailable or inconvenient in some cases.

\section{Method}
 
\begin{figure}[!t]   
	\centering
	\includegraphics[width=\linewidth]{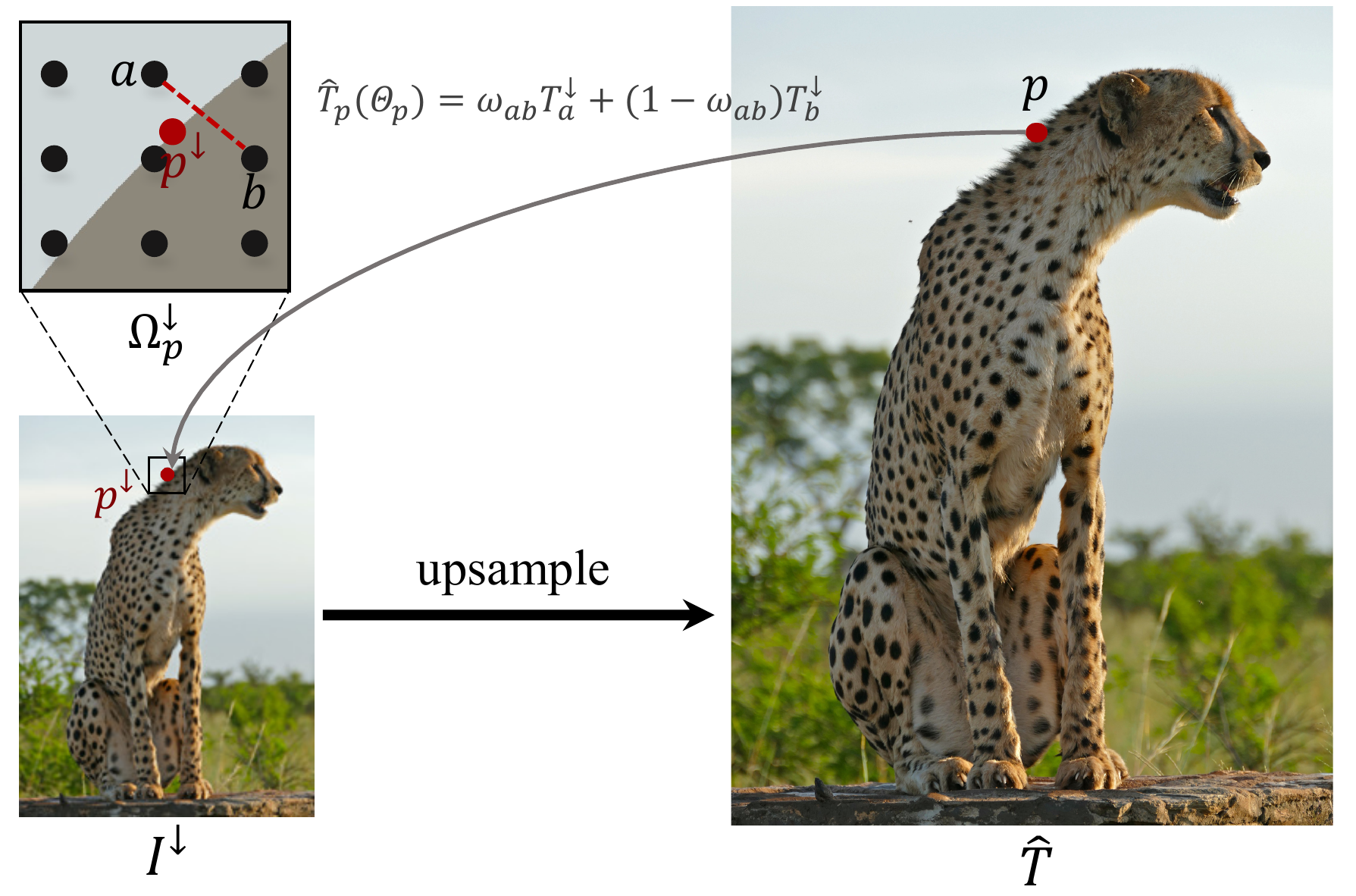}
	\caption{ The proposed linear representation of high-resolution images. Each pixel $p$ in the high-resolution image is represented as the linear interpolation of two pixels $(a,b)$ in the low-resolution image, with $(a,b)$ and the interpolation weight optimized for minimizing the representation error.}
	\label{fig:interpolation}
\end{figure}

Given an image operator $f$ and a high-resolution input image $I$ with RGB colors, our goal is to obtain an approximation $\hat{T}$ of the original output $T=f(I)$. With the guided upsampling method, we first apply $f$ to a downsampled image $\Ilow$, and then upsample the result $\Tlow$to the high-resolution output $\hat{T}$ with the guidance of $I$. We need to optimize the downsampling and upsampling processes in order to minimize the difference between $\hat{T}$ and $T$. Note that the same as previous universal guided upsampling methods~\cite{kopf2007joint,chen2016bilateral}, we also treat $f$ as a black-box operator that is scale-invariant.

\begin{figure*}[ht] 
	\centering 
	\subfigure[Source (560$\times$560)]{
		\includegraphics[width=0.185\linewidth]{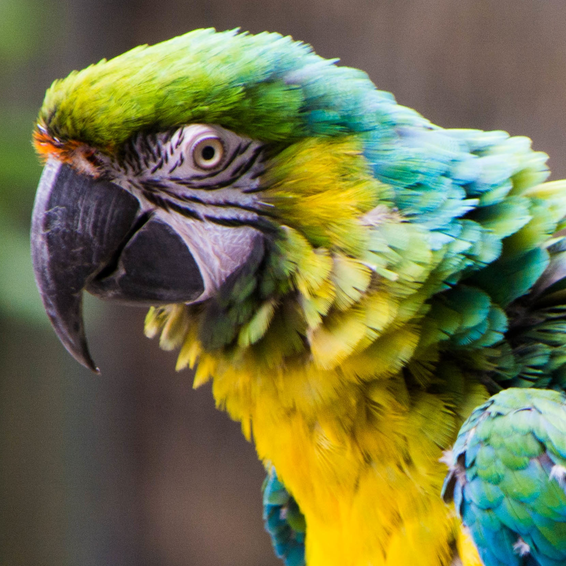}
	} 
	\subfigure[JBU ($8\times$)]{
		\includegraphics[width=0.185\linewidth]{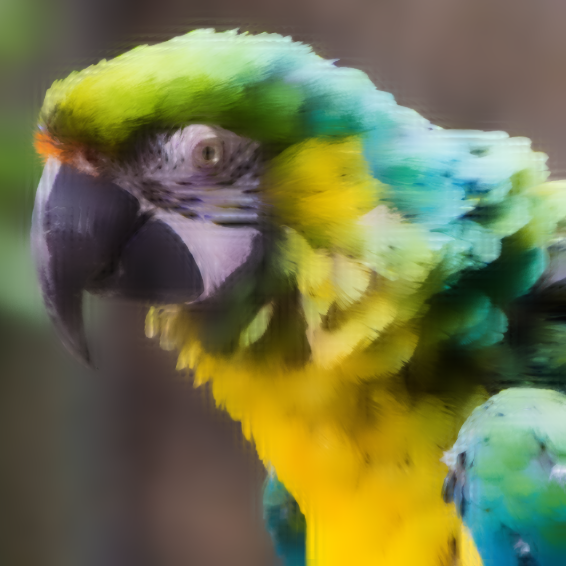} 
	}
	\subfigure[GLU$^*$ ($16\times$)]{
		\includegraphics[width=0.185\linewidth]{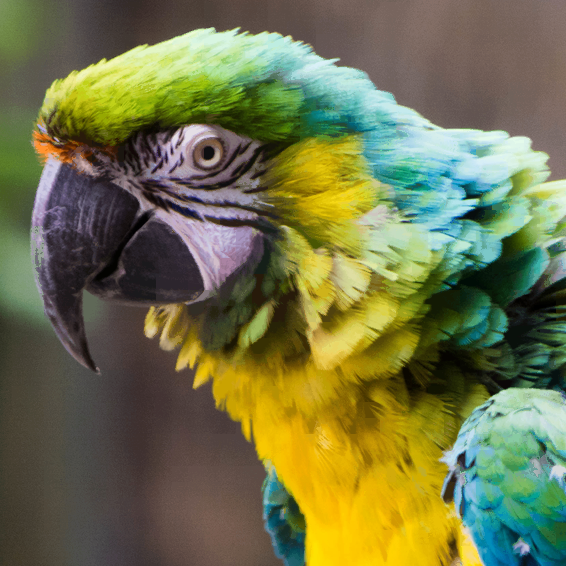} 
	} 
	\subfigure[GLU$^*$ ($32\times$)]{
		\includegraphics[width=0.185\linewidth]{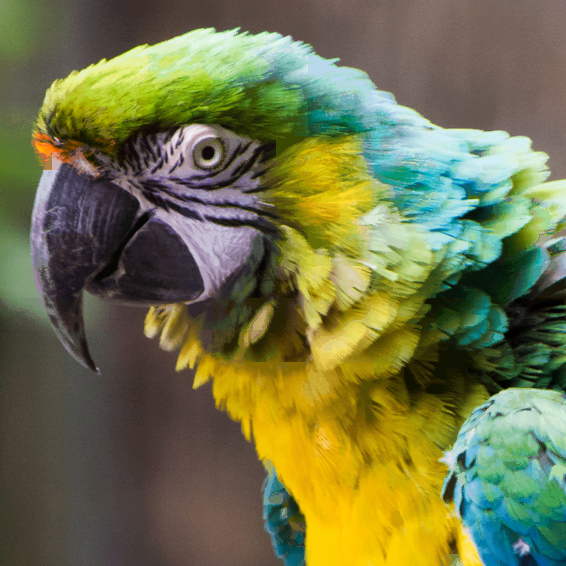} 
	} 
	\subfigure[GLU ($32\times$)]{
		\includegraphics[width=0.185\linewidth]{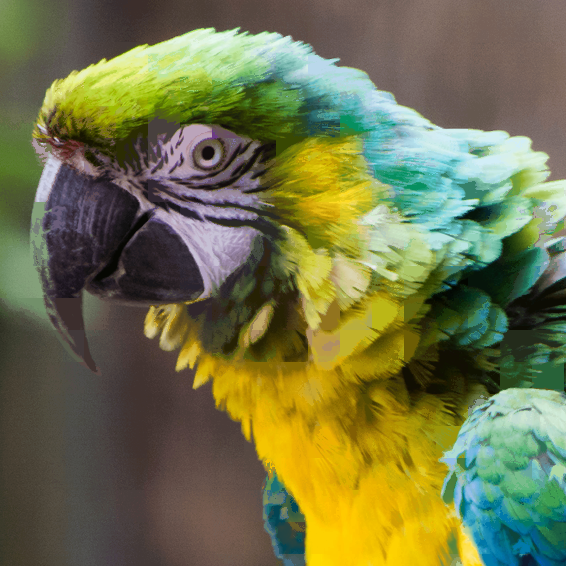} 
	} 
	\vspace{-3mm}
	\caption{Guided self-upsampling with JBU and our method. The input source image is downscaled and then upsampled to the original resolution with the guidance of itself. GLU$^*$ is our method as described in Section \ref{sec:glu}, GLU is the accelerated version introduced in Section \ref{sec:glu-fast}. They both can well recover the original image for large ratios as $32\times$. As the comparison, JBU produces obvious blur even for smaller ratios as $8\times$. }
	\label{fig:withjbu} 
\end{figure*}

\subsection{Guided Linear Upsampling}
\label{sec:glu}

We first assume that the downsampled input image $\Ilow$ is given, or produced with regular grid downsampling by default. As illustrated in \reffig{fig:interpolation}, in order to optimize the upsampling, our basic assumption is that each pixel $p$ of the high-resolution target image $T$ can be well approximated by the linear interpolation of a pair of low-resolution pixels $(a,b)$ as follows:
\begin{equation}
\label{eq:interpT}
	\hat{T}_p=\omega_{ab} \Tlow_a+(1-\omega_{ab}) \Tlow_b \;\; s.t.\;\; a,b\in\Omegaplow
\end{equation}
where $\omega_{ab}$ is the weighting function, $p^\downscale$ is the downscaled coordinates of $p$, and $\Omegaplow$ is a small neighborhood of $p^\downscale$. $\hat{T}_p$ is an estimate of the original output $T_p$. The above assumption can actually be derived from the well-known \emph{color line model}~\cite{levin2007closed}, which we will explain in \onerevise{Section} \ref{sec:derivation}. \refeq{eq:interpT} contains three parameters $a,b,\omega_{ab}$, which need to be optimized in order to minimize the upsampling error. Denote the parameters of pixel $p$ by $\Theta_p=\{a,b,\omega_{ab}\}$, and then $\Theta$ is a $3\times H \times W$ tensor containing the parameters of all pixels of $T$. Given $\Theta$, the corresponding high-resolution output $\hat{T}(\Theta)$ can be easily computed with \refeq{eq:interpT}. 

\onerevise{The same as previous \emph{local color transformation} methods~\cite{levin2007closed, he2012guided, chen2016bilateral}, we also assume that the target image can be locally represented as the affine transformation of the source image. As will be explained in Section \ref{sec:derivation}, in this case the source and target images can be optimally upsampled with the same set of parameters.} In other words, if $\Theta$ is optimal for the source image, then it should be optimal for the target image as well. 
%
%To determine $\Theta$, we further assume that the source and target images can be optimally upsampled with the same set of parameters. In other words, if $\Theta$ is optimal for the source image, then it should be optimal for the target image as well. Actually, this assumption can be derived from previous \emph{local color transformation} methods~\cite{levin2007closed, he2012guided, chen2016bilateral}, as will be explained in section \ref{sec:derivation}. 
%Therefore, we can solve the optimal parameters for the guidance image $I$, and then apply the same parameters to upsample the target image. The principles behind this will be discussed in section \ref{sec:derivation}. 
Therefore, the optimal parameters $\Theta$  can be solved w.r.t only the source image in order to minimize its upsampling error:
%with respect to the linear interpolation in \refeq{eq:interpT}:
\begin{equation}
	\Theta=\mathop{\arg\min}\limits_{\Theta} \parallel \hat{I}(\Theta) - I \parallel
\end{equation}
in which $\hat{I}(\Theta)$ is the upsampled source image with the given parameters $\Theta$. We assume that $\Theta_p$ of each pixel is independent of each other, so the above equation can be solved for each pixel as
\begin{equation}
	\Theta_p=\mathop{\arg\min}\limits_{\Theta_p} \parallel\omega_{ab} \Ilow_a + (1-\omega_{ab})\Ilow_b-I_p\parallel
\label{eq:opt-pixel}
\end{equation}
which is a combinatorial optimization problem that is usually difficult to solve. Fortunately, in our case $\Omegaplow$ is a small neighborhood (a $3\times 3$ window in our experiments), so it is easy to enumerate all possible pixel pairs. For each selected pixel pair $(a,b)$, the optimal weighting parameter should be
\begin{equation}
	\omega_{ab} = \frac{( I_p - \Ilow_b)(\Ilow_a - \Ilow_b)}{\parallel \Ilow_a - \Ilow_b\parallel^2 + \varepsilon}
	\label{eq:wab}
\end{equation}
which makes the interpolation result $\hat{I}_p$ the projection of $I_p$ on the color line determined by $\Ilow_a$ and $\Ilow_b$, just as previous sampling-based matting methods~\cite{wang2007optimized}. \revised{$\varepsilon$ is a small constant ($10^{-3}$ in our implementation) to avoid dividing by zero in flat patches.}

\reffig{fig:withjbu} demonstrates an example to upsample an image from its downscaled counterpart. The above simple method achieves surprisingly good results. Even for large ratios such as $32\times$, details of the original image can be reconstructed almost perfectly. As a comparison, the result of JBU is obviously blurred even for smaller ratios. Note that for the methods based on the local color transformation~\cite{he2012guided,chen2016bilateral}, the above task is trivial because an identity transformation would be learned if the source and target images are the same.

\begin{figure}[ht]
	\centering 
	\subfigure[Source (112$\times$168)]{
		\includegraphics[width=0.31\linewidth]{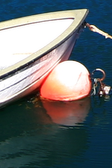}
	} \hspace{-1.5mm}
	\subfigure[GNU]{
		\includegraphics[width=0.31\linewidth]{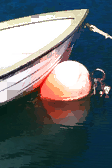}
	} \hspace{-1.5mm}
	\subfigure[GLU]{
		\includegraphics[width=0.31\linewidth]{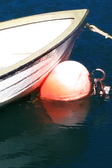} 
	}
	%	} \hspace{-1.5mm}
%	\subfigure[GLU]{
	%		\includegraphics[width=0.23\linewidth]{figure/fig4/4glu.png} 
	%	} 
\vspace{-3mm}
\caption{Comparison of GLU and GNU for upsampling an image patch from its $8\times$ downsampling. GNU cannot represent the smooth variations of the original image, and thus produces obvious artifacts. }
\label{fig:reconstruct} 
\end{figure}

\subsection{Efficient Computation}
\label{sec:glu-fast}
The complexity of the above method is quadratic to the number of pixels in $\Omegaplow$. For a typical $3\times 3$ window, it needs to check 36 pairs of pixels in order to minimize \refeq{eq:opt-pixel}. For high-resolution images, this still requires a large amount of computation, so we propose the following improvements for better efficiency.

Firstly, we find that it is not necessary to enumerate all pixel pairs $(a,b)\in\Omegaplow$ in order to optimize \refeq{eq:opt-pixel}. Instead, we can first fix $a$ as the pixel with the most similar color to $I_p$, and then optimize only $b$ and $\omega_{ab}$ with respect to \refeq{eq:opt-pixel}. In this way, the complexity can be reduced to be linear with $|\Omegaplow|$. Since $a$ is close to $I_p$ in the color space, the approximation error should be small for the projection of $I_p$ on the color line.

Secondly, it is easy to see that if $I_p$ is on the color line determined by $\Ilow_a$ and $\Ilow_b$, the interpolation weight $\omega_{ab}$ as in \refeq{eq:wab} reduces to
\begin{equation}
	\omega_{ab}=\frac{\parallel I_p-\Ilow_b \parallel}{\parallel I_p-\Ilow_a \parallel + \parallel I_p-\Ilow_b \parallel + \varepsilon}
\label{eq:wab-fast}
\end{equation}
which can be computed more efficiently and the results are guaranteed to be in $[0,\,1]$. Since the color lines not crossing $I_p$ are less likely to be selected, the above approximation has little impact on the quality of our method.

As shown in \reffig{fig:withjbu}, the above accelerations would not introduce noticeable differences compared to our original method, but the complexity is much lower. Therefore, in the following we will use the accelerated method by default.

Our final upsampling method is as described in the Algorithm \ref{alg:alg1}. It is very simple and efficient. $\Omegaplow$ is typically chosen as a $3\times 3$ window, so for each pixel, only 9 pixel pairs need to be checked. Note that if we fix $\omega_{ab}$ to 1, then the optimization in line 3 is not necessary, and $\hat{T}_p$ would be equal to $\Tlow_a$. We call this special case of our method as \emph{Guided Nearest Upsampling} (GNU). As shown in \reffig{fig:reconstruct}, GNU lacks the ability to recover the ramp edges and smooth variations of natural images, thus producing blocky effects and false contours, which can be effectively eliminated by using GLU.

\begin{algorithm}
	\caption{Efficient Guided Linear Upsampling.}
	\label{alg:alg1}
	\raggedright
	\KwIn{High-res source image $I$, low-res source image $\Ilow$ and corresponding target image $\Tlow$.}
	\KwOut{High-res target image $\hat{T}$.}
	\BlankLine
	
	\For{each pixel $p\in I$}{
		Find $a$ as the pixel in $\Omega_p^\downscale$ with the most similar color to $I_p$ \;
		Fix $a$ and optimize $b,\omega_{ab}$ with Eq. (\ref{eq:opt-pixel})(\ref{eq:wab-fast})\;
		Compute $\hat{T}_p$ as \refeq{eq:interpT}\;
	}
\end{algorithm}

\begin{figure*}
	\centering 
	\subfigure[Source  (284$\times$202)]{
		\includegraphics[width=0.1865\linewidth]{figure/fig7-od/1-gt.pdf}
	} 
	\subfigure[$16\times$ downsampling]{
		\includegraphics[width=0.1895\linewidth]{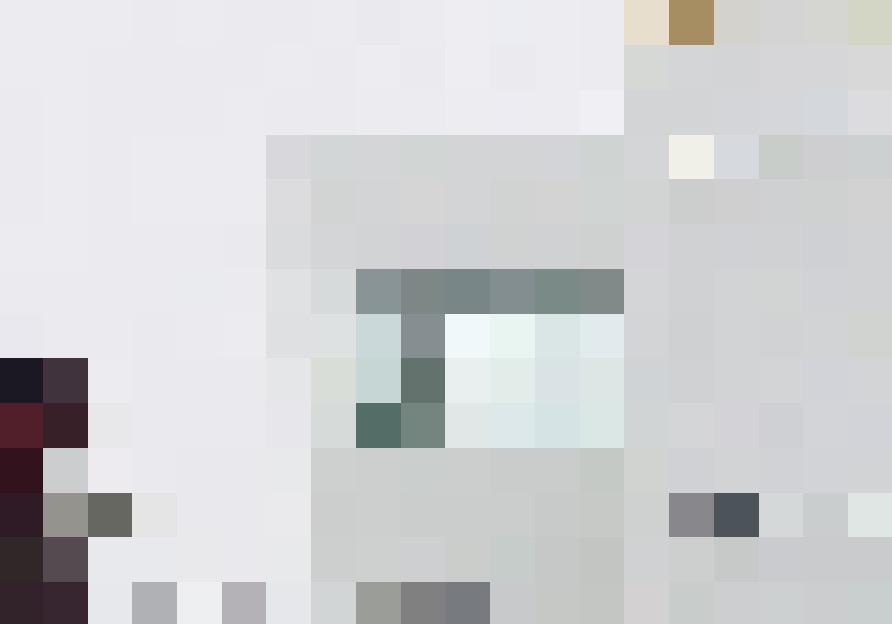}
	} 
	\subfigure[GLU$^-$]{ 
		\includegraphics[width=0.1865\linewidth]{figure/fig7-od/1-wo.pdf}
	}
	\subfigure[\modified{Optimized downsampling}]{
		\includegraphics[width=0.1895\linewidth]{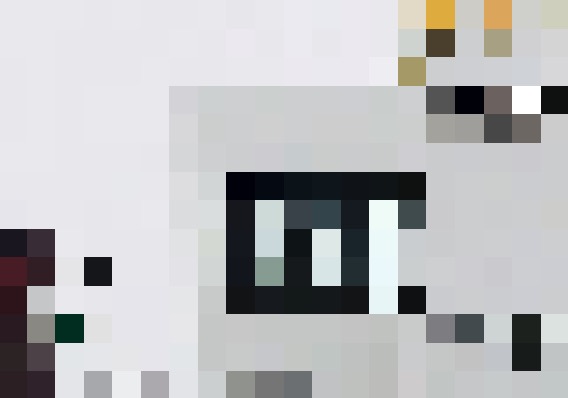} 
	} 
	\subfigure[GLU]{
		\includegraphics[width=0.1865\linewidth]{figure/fig7-od/1-glu.pdf} 
	} 
        \vspace{-3mm}
	\caption{Demonstration of downsample optimization. (a) The input image with some thin structures. (b) Most thin structures are lost with $16{\times}$ default downsampling. (c) The result upsampled from (b), the thin structures cannot be recovered. (d) Optimized $16{\times}$ downsampled image. (e) The result upsampled from (d), the thin structures are well recovered.} 
	\label{fig:downsample} 
\end{figure*}

\begin{figure*} 
	\centering
	\includegraphics[width=\linewidth]{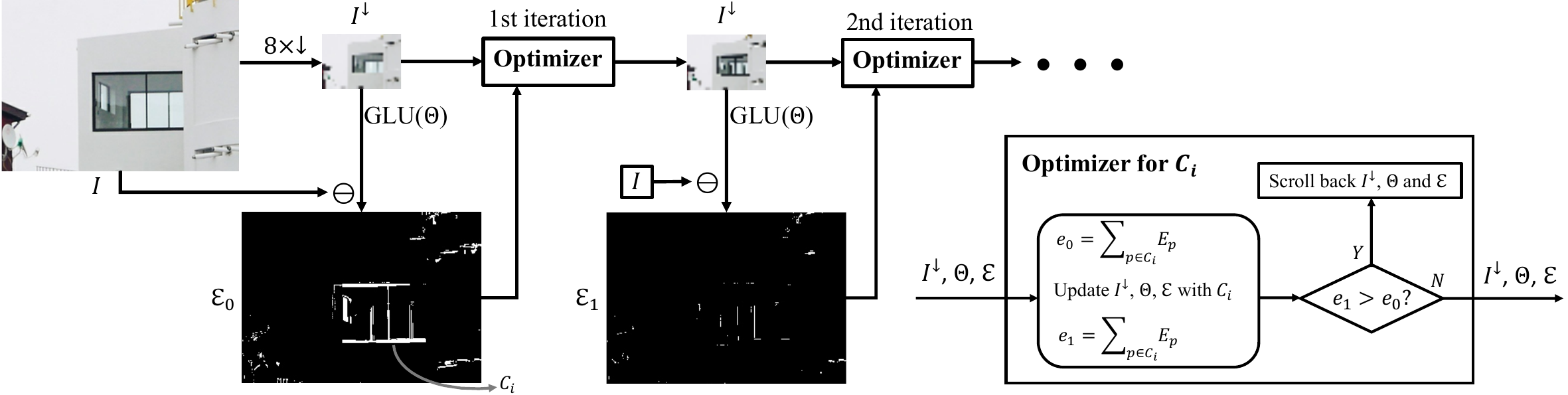}
	\caption{Illustration of the proposed downsample optimization method. For the input high-resolution image $I$, we initialize $I^{\downarrow}$ with regular grid downsampling, and then iteratively update $I^{\downarrow}$ by trying to add large-error pixels to it for minimizing the total upsampling error.}
	\label{fig:downflow}
\end{figure*}

\subsection{Downsample Optimization}
\label{sec:downsample}

For large downsampling ratio, isolated thin structures and small regions may be completely lost due to regular grid downsampling. In this case, it would be impossible for the upsampling process to recover the original content. \reffig{fig:downsample} demonstrates such a situation. Although downsampling optimization has been extensively studied, previous works mainly aim to avoid aliasing artifacts~\cite{kopf2013content,oeztireli2015perceptually,weber2016rapid}, which is different from our goal. Some super-resolution methods~\cite{kim2018task,sun2020learned,xiao2020invertible} also jointly optimize their downscaling and upscaling processes, which however, are not well suited for the proposed GLU upsampler. %We thus propose an effective downsampling optimization method.

Given the GLU upsampler $\Psi(\Ilow, \Theta)$, we can formulate the downsampling process as an optimization problem aiming to minimize the self-upsampling error of the source image. In practice since $\Theta$ is unknown, the downsampling and upsampling need to be jointly optimized as
\begin{equation}
	\Ilow, \Theta=\mathop{\arg\min}\limits_{\Ilow, \Theta} \parallel I-\Psi(\Ilow, \Theta) \parallel
	\label{eq:opt-downsample}
\end{equation}
with each pixel of $\Ilow$ from exactly one pixel of $I$. Note that this is different from previous downsampling optimization methods, in which each pixel of $\Ilow$ is usually filtered from multiple pixels of $I$ in order to reduce aliasing artifacts. For our method, the filtering in downsampling may significantly blur the upsampled image because it would shrink the endpoints of color lines, which is detrimental to image details.

\refeq{eq:opt-downsample} can be solved by iteratively optimizing $\Ilow$ and $\Theta$. Given $\Ilow$, the upsampling parameters $\Theta$ can be solved as in Algorithm \ref{alg:alg1}. To optimize the downsampling, we first compute the pixel-wise error map $E, E_p=\parallel I_p-\hat{I}_p\parallel$. Obviously, the pixels with large error must be those that cannot be well represented by $\Ilow$, and thus need to be added to $\Ilow$ by replacing some existing pixels. \onerevise{Note} that since each pixel in $\Ilow$ may be used to interpolate multiple pixels of $I$, the above operation may not reduce the total error. Therefore, we adopt a trial-and-error approach, and if replacing some pixels in $\Ilow$ does not reduce the total error, the replaced pixels would be rolled back.

\reffig{fig:downflow} illustrates the procedure of our method, more details are described in Algorithm \ref{alg:joint}. The trial-and-error procedure is executed for each connected region $C_i$ of pixels with large error ($\mathcal{E}$). The pixels with large errors are tried to be added to the downsampled image, and the operation would be accepted if it can reduce the total error, otherwise it would be unrolled. For multiple high-resolution pixels $[q^{_\uparrow}]$ mapped to the same low-resolution pixel location $q\in\Ilow$, the one with the largest error would be selected to replace the original color of $q$.

As shown in \reffig{fig:downsample}, the above method can effectively prevent the missing of thin structures and small regions. In most cases, it requires only 1 or 2 iterations to converge, and after the initialization, only pixels with large errors are involved for further processing, so only a little more computation is required.

\newcommand{\Ilown}[1]{I^{_{\downarrow #1}}}

\begin{algorithm}
\caption{Joint Optimization of Down- and Upsampling.}
\label{alg:joint}
\raggedright
\KwIn{High-res source image $I$, the error threshold $\tau$, the maximum iterations $N$.}
\KwOut{Optimized low-res image $\Ilow$ and upsampling parameters $\Theta$.}
\BlankLine

Initialize $\Ilow$ with regular grid downsampling\;
Initialize $\Theta$ from $I,\; \Ilow$ with Algorithm 1\; 
Compute initial error map $E$\;
\For{n = 1, ..., N}{
	Find the set of pixels with large error: $\mathcal{E}=\{p\,|\,E_p>\tau\}$ \;
	\If{$\mathcal{E}$=$\emptyset$}{Break} 
	Cluster $\mathcal{E}$ as connected components $C_1, \cdots, C_M$\;	
	\For{$i=1,\cdots,M$}{
		Backup $\Theta, \Ilow, E$ for scroll back\;
		Compute $e^0=\sum_{p\in C_i} E_p$\;
		$Q=\{q\,|\,q\in\Ilow\,\&\,[q^{_\uparrow}]\bigcap C_i\neq\emptyset$\}\;
		\For{$q\in Q$}{
				Update $\Ilow_q$ with $I_p, \; p=\mathop{\arg\max}\limits_{p\in [q^{_\uparrow}]\bigcap C_i} E_p$
		}	
		\For{$p\in C_i$}{
			Update $\Theta_p$ as in Algorithm 1\;
			Update $E_p$\ with updated $\Theta_p$\;
		}
		Compute $e^1=\sum_{p\in C_i} E_p$\;
		\If{$e^1>e^0$}{
			Scroll back updated regions of $\Theta,\Ilow$ and $e$\;
		}
	}
}

\end{algorithm}

\section{Analysis}
\label{sec:analysis2}

An ideal guided upsampling method should be able to preserve the detail effects of the target image while avoiding artifacts such as bleeding and blurring. In the following we will analyze the capabilities of our method and show how it relates to previous methods.

\subsection{Theoretical Derivation}
\label{sec:derivation}

The proposed upsampling method in Section \ref{sec:glu} can be derived from the color line model~\cite{levin2007closed} and local color transformation methods~\cite{levin2007closed, he2012guided, chen2016bilateral}. 

The color line model tells us that the colors of pixels in a small patch should be roughly on the same line in the color space. Therefore, the color of each pixel in the patch must be well approximated by the linear interpolation of the two endpoints $[a, b]$ of the color line. After downsampling, it can be expected that $[a, b]$ still can be well represented by two pixels in the downsampled patch, because of the information redundancy in the high-resolution image. As a result, each pixel color in the original patch can also be linearly interpolated by two pixels in the downsampled patch, as in \refeq{eq:interpT}. 

The local color transformation methods assume that the output image can be locally represented as the  affine transformation of the input image, i.e. $T_p=A_p I_p$, where $A_p$ is an affine transformation that varies smoothly over the image space. 
%\revised{In addition, since the image operator is considered to be scale-invariant, $A_p$ should keep the same for different scales, so $\Tlow_p=A_p \Ilow_p$.} 
\onerevise{In addition, we require the operator to be approximately scale-invariant: $\Tlow_p=A_p \Ilow_p$.}
Therefore, if using $\Theta_p$ can linearly interpolate $I_p$, i.e.
\begin{equation}
	I_p=\omega_{ab} \Ilow_a+(1-\omega_{ab}) \Ilow_b
\end{equation} 
then it immediately \onerevise{follows that} 
\begin{equation}
	T_p= \omega_{ab}A_p\Ilow_a+(1-\omega_{ab})A_p\Ilow_b = \omega_{ab}\Tlow_a+(1-\omega_{ab})\Tlow_b
\end{equation}
which means that using $\Theta_p$ also can linearly interpolate $T_p$, as we have assumed in Section \ref{sec:glu}.

\subsection{Edge Recovery}
\label{sec:edge}

Typical image edges can be classified into three types: step edge, ramp edge, and roof edge~\cite{koschan2005detection,yin2019side}. For natural images, most edges should be ramp edges connecting two regions. Obviously, the transition effects of ramp edges can be well represented by linear interpolation of the two region colors. Therefore, by interpolating only two pixels, GLU can recover the edges of the original image very well. As a comparison, using GNU can recover only the step edges, and thus would introduce significant artifacts, as is shown in \reffig{fig:reconstruct}.

A natural question is whether we can achieve further improvements by interpolating more pixels. Indeed, \refeq{eq:interpT} can be more generally expressed as
 \begin{equation}
 	\hat{T}_p=\sum\limits_{q\in\Omegaplow} \omega_q\;\Tlow_q 
 	\label{eq:general}
 \end{equation}
with $\omega_q$ as the normalized weights. Interestingly, this is exactly the form of JBU. However, in JBU $\omega_q$ is not optimized, and the filtering effect would result in blur and edge reversal artifacts~\cite{he2012guided}. It is easy to see that when $\sigma_d\to \infty$ and $\sigma_r\to 0$, JBU will reduce to GNU. However, in practice this is hard to achieve due to the numerical problems of the \emph{exp} weighting function. By decreasing $\sigma_r$, the blurring artifacts of JBU can be reduced, but may lead to aliasing artifacts as GNU. Therefore, in this sense both GLU and GNU can be taken as special cases of JBU with optimized weights.

Although not tested, we do not see the need to take more pixels for interpolation. Involving more pixels not only makes the optimization more difficult, but may also lead to overfitting and extrapolation, both of which can reduce the result quality. 

\begin{figure}[ht]   
	\centering
	\includegraphics[width=\linewidth]{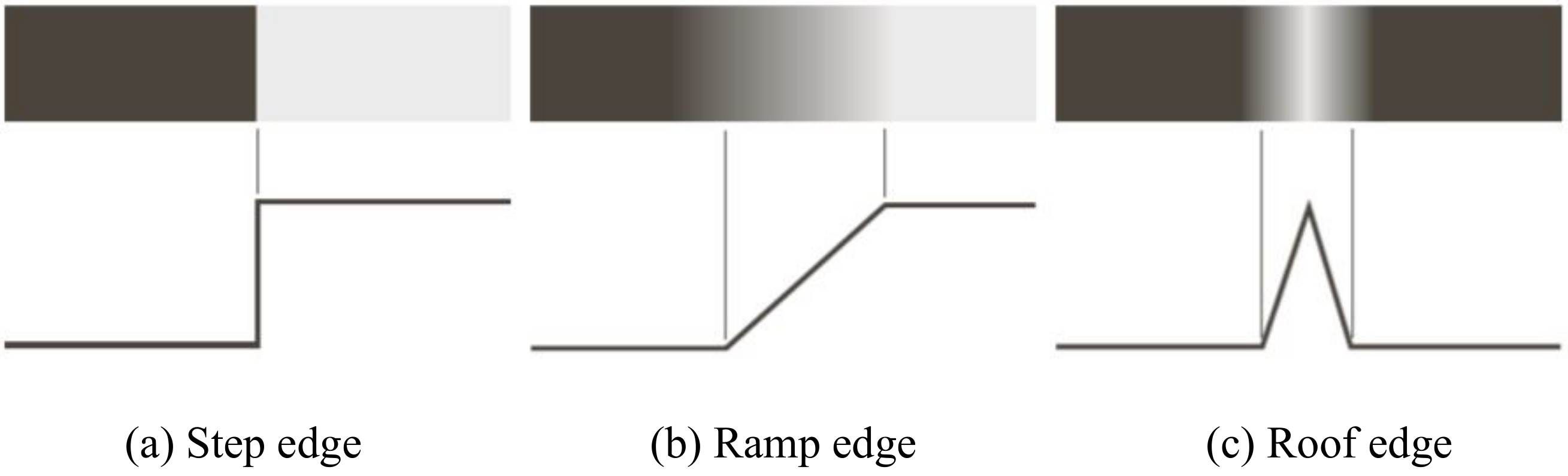}
	\caption{The three types of edges in natural images.}
	\label{fig:edges}
\end{figure}

\subsection{Detail Preservation}
\label{sec:details}

As discussed in Section \ref{sec:derivation}, our method implicitly takes advantage of the local color transformation for transferring the upsampling parameters. However, it should be noted that unlike previous approaches such as guided filter~\cite{he2012guided} and BGU~\cite{chen2016bilateral}, our method does not require the transformations to be smooth in either image space or bilateral space. Therefore, it can better preserve the detail effects of the target image while avoiding the bleeding artifacts caused by over-smoothing.

% One may concern that the lack of smooth constraint will introduce artifacts. Although keeping smoothness is important for most image processing operators, we find that our method with each pixel independently solved also works well in most situations.
\onerevise{One potential issue with our method is the lack of an explicit smoothness constraint.} Although preserving smoothness is important for most image processing operators, we find that our method \onerevise{which operates on each pixel independently} also works well in most situations. This is mainly because the linear interpolation model can well approximate the appearance of the original source image, which serves as a smooth guidance map that can suppress unsmooth artifacts if the target image has similar local affinities as the source image. However, if the pixel affinities of the source and target images are significantly different (e.g., when new edges are introduced in the target image), unsmooth artifacts may be produced. Actually, this is the main limitation of our method, which we will discuss further in \onerevise{Section} \ref{sec:limitations}.

\section{Experiments}
\label{sec:experiments}

In experiments we evaluate the proposed method in various image processing applications, and compare it qualitatively and quantitatively with  previous methods. We also demonstrate the advantages of our method for interactive image editing and real-time video processing, and reveal its limitations for more diverse applications.

%We illustrate qualitative results comparison in Sec.~\ref{sec:5.1}, and quantitative results comparison in Sec.~\ref{sec:5.2}. Results are best viewed electronically on a high resolution monitor.

\begin{table*}[!ht]
\scriptsize  %fontsize
	\begin{minipage}[!t]{\linewidth}
		\caption{Comparisons of different methods with PSNR scores. The low-resolution target is produced from the low-resolution source using  the image operator, except the applications with $^\dagger$, for which the low-resolution target is obtained by downsampling the reference image. }
            \vspace{-3mm}
		\renewcommand{\arraystretch}{1.25}%
		\label{tab:psnr}
		\centering
		%  	\scriptsize%
% 		\setlength{\tabcolsep}{2.9mm}{} 
		\resizebox{\linewidth}{!}{
			\begin{tabular}{ccccccccccccccc}
				\toprule
				\multirow{2}*{PSNR$\uparrow$}
				& \multicolumn{2}{c}{alpha matting} &  \multicolumn{2}{c}{colorization} & \multicolumn{2}{c}{unsharp mask} & \multicolumn{2}{c}{$L_0$-smoothing}  & \multicolumn{2}{c}{dehazing} &\multicolumn{2}{c}{laplacian filter} & \multicolumn{2}{c}{unsharp mask$^\dagger$} \\ 
				\cmidrule(r){2-3}\cmidrule(r){4-5}\cmidrule(r){6-7}\cmidrule(r){8-9}\cmidrule(r){10-11}\cmidrule(r){12-13}\cmidrule(r){14-15}
				& $8\times$ &   $16\times$  &   $8\times$ &   $16\times$ &   $8\times$  
				& $16\times$ &   $8\times$  &   $16\times$ &   $8\times$ &   $16\times$ 
				& $8\times$ &   $16\times$  &   $8\times$ &   $16\times$\\
				\midrule
				JBU  &25.6	& 22.9	&20.9	 &20.1	& 18.2	& 16.9	 & 22.3&20.1	& 25.9	 & 22.3  & 15.6&13.8  &19.0&17.8\\
				BGU-fast &21.4 & 22.2 &28.5 & 27.8 & 23.8 & \textbf{23.2} & 22.8 & 22.2 & 21.1 & 17.7 & 21.8  & 21.3 &25.1&24.9\\
				BGU  & 28.3  & 25.8	 & 30.7	 & 28.8	 & 23.5  & 22.4	 &  27.0 & 25.4 & 26.8 & 23.4 & \textbf{23.7} & 22.3 &25.4&25.0\\
				\midrule
                GLU$^-$ &  31.4	&  28.9 	&  29.7	&  27.7 &	 23.6 &22.3	 & 23.6 & 24.5	&  27.6	& 24.1	&  20.5&17.2 &25.2&24.0\\
				% GLU$^-$ &  31.37	&  28.90 	&  29.73	&  27.77 &	 23.57 &22.30	 & 28.00 & 26.22	&  27.60	&  24.13	&  29.50	&  27.91 \\

				GLU  & \textbf{ 31.5}	& \textbf{ 29.1}	& \textbf{ 31.3}	&\textbf{ 29.6}	&\textbf{ 24.0}	& 22.4	& \textbf{ 28.8} &	 \textbf{27.1} &\textbf{ 27.6}	&\textbf{ 24.1}	&23.1&\textbf{24.5} &\textbf{25.9}&\textbf{25.2}\\

				\bottomrule
		\end{tabular}}
	\end{minipage}
	\\[12pt]
	\begin{minipage}[!t]{\linewidth}
		\caption{\modified{Comparisons of different methods with SSIM scores.}}
            \vspace{-3mm}
		\renewcommand{\arraystretch}{1.25}
		\label{tab:ssim}
		%  	\scriptsize%
		\centering
		\resizebox{\linewidth}{!}{
			\begin{tabular}{ccccccccccccccc}
				\toprule
				\multirow{2}*{SSIM$\uparrow$}
				& \multicolumn{2}{c}{alpha matting} &  \multicolumn{2}{c}{colorization} & \multicolumn{2}{c}{unsharp mask} & \multicolumn{2}{c}{$L_0$-smoothing}  & \multicolumn{2}{c}{dehazing} &\multicolumn{2}{c}{laplacian filter} & \multicolumn{2}{c}{unsharp mask$^\dagger$}\\ 
				\cmidrule(r){2-3}\cmidrule(r){4-5}\cmidrule(r){6-7}\cmidrule(r){8-9}\cmidrule(r){10-11}\cmidrule(r){12-13}\cmidrule(r){14-15}
				& $8\times$ &   $16\times$  &   $8\times$ &   $16\times$ &   $8\times$  
				& $16\times$ &   $8\times$  &   $16\times$ &   $8\times$ &   $16\times$
				& $8\times$ &   $16\times$ &$8\times$  &$16\times$\\
				\midrule
				JBU  &   0.93	&   0.91	&   0.60	&   0.56	&   0.40   &	  0.36   &	   0.80   &   0.78	&   0.91	& 0.88	&   0.32&0.27 &0.41&0.37\\
				BGU-fast & 0.71 & 0.64 & 1.00 & 1.00 & 0.85 & 0.84 & 0.83 & 0.82 & 0.79 & 0.72 & 0.82 & 0.81 &0.77&0.75\\
				BGU  &   0.86	&   0.78	& \textbf{1.00}	& \textbf{1.00}	& \textbf{0.88}   &	\textbf{0.85}   &	   0.88   &  0.84	&   0.90	&   0.85	&  \textbf{ 0.88}	&  \textbf{ 0.84} &0.79&0.77\\
				\midrule
                    GLU$^-$ & \textbf{0.96}	& \textbf{0.94}	&   0.97	&   0.97	&   0.86   &	  0.83   &	   0.86   &  0.83	&   \textbf{0.94}	&   \textbf{0.89}	&   0.80&0.68 &0.82&0.79\\
				% GLU$^-$ & \textbf{0.96}	& \textbf{0.94}	&   0.97	&   0.97	&   0.86   &	  0.83   &	   \textbf{0.89}   &  \textbf{0.85}	&   \textbf{0.94}	&   \textbf{0.89}	&   0.94	&   \textbf{0.92}\\
				GLU  & \textbf{0.96}	& \textbf{0.94}	&   0.99	&   0.99	&   0.87   &	  0.83   &	 \textbf{0.89}   &  \textbf{0.85}	& \textbf{0.94}	&   \textbf{0.89}	& 0.85&0.80 &\textbf{0.83}&\textbf{0.81}\\
				\bottomrule
		\end{tabular}}
	\end{minipage}
\end{table*}

\subsection{Comparisons}
\label{sec:5.1}
% We apply our method to 116 images from XX Datasets on following applications: 
% $\bullet$ \textbf{Reconstructing} 
%Our technique can be used to upsample an low resolution image with guidance of the high resolution one. Operators whose discontinuities typically correspond to edges in an image have desirable properties of our technique. We validate the effectiveness of \textit{guided linear upsampling} in multiple application scenarios, and assume the results quality by computing their PSNR and SSIM with respect to ground truth for the entire experiments. Qualitative results of applications examples are compared in \reffig{fig:results}. 
%The operators we demonstrate are:

%The same as previous guided upsampling methods~\cite{chen2016bilateral}, our method also requires the discontinuities appear at only source image edges, i.e. the image operator would not introduce new edges or spatial warps. 
For quantitative evaluations we tested our method with the following applications and datasets:

$\bullet$ \textbf{Alpha Matting} with the method of \cite{chen2013knn}. The dataset is from \cite{rhemann2009perceptually}, which consists of 27 high-resolution images with a size of about $6-8$M pixels.
%and the reference alpha matte images are produced by \cite{chen2013knn}. %Each image contains 6M pixels.

$\bullet$ \textbf{Colorization} with the method of \cite{levin2004colorization}. The dataset is from the high-quality 2K images for super-resolution~\cite{Agustsson_2017_CVPR_Workshops}. We use all 100 images in the validation set. The source grayscale images are produced by graying the original RGB images and then converted to 3 channels simply by replicating the channel. The required seed constraints~\cite{levin2004colorization} are sparsely sampled from the original color images. 

$\bullet$ \textbf{Dehazing} with the method of \cite{li2017aod}. The dataset is from NTIRE-19 benchmark \cite{ancuti2019ntire}, which includes 55 real hazy images with $\sim$2M pixels.

$\bullet$ \textbf{Unsharp Masking} for enhancing image details by the method of \cite{ngo2020nonlinear}. The source images are the same as \emph{colorization}.

$\bullet$ \textbf{$L_0$ Smoothing} using the method of \cite{xu2011image}. The dataset is the same as the \emph{colorization}.

$\bullet$ \textbf{Laplacian Filtering} for enhancing image details with the method of \cite{aubry2014fast}. The dataset is the same as \emph{colorization}.

%\sout{$\bullet$ \textbf{Optical Flow} with the RAFT flow method \cite{teed2020raft} and test images from the MPI Sintel dataset~\cite{Butler:ECCV:2012}, whose image size is small ($1024\times 468$), so we test with only smaller downsampling ratios of $4\times$ and $8\times$.}
 
%Darkness lighting: corrects contrast of under-/over-exposed images with low contrast. We use the algorithm of \cite{cai2018learning} for evaluation, and select the first 20 pairs of images from their dataset as the validation set. %VV: https://sites.google.com/site/vonikakis/datasets

% \modified{For applications except optical flow, the low-resolution target images are produced from the corresponding low-resolution source images with the chosen image operators, and then upsampled to get the output full-res target. The reference output is obtained with the image operator in full resolution. For optical flow, since the flow offset computed in downscaled images is largely different from the reference, we produce the low-resolution target by downsampling the reference, which forces the operator to be scale-invariant and should be better to see the net effect of upsampling methods.}

%For each test example, the low-resolution target image is first downscaled from the high-resolution reference, and then upsampled with different methods for the evaluation. 

%set $\Omegaplow$ as a $3\times 3$ window and the downsampling error thresold $\tau$ as $30/255$. 

%It should be mentioned that, for different size of inputs..

\begin{figure*}[!t]   
	\centering
	\includegraphics[width=\linewidth]{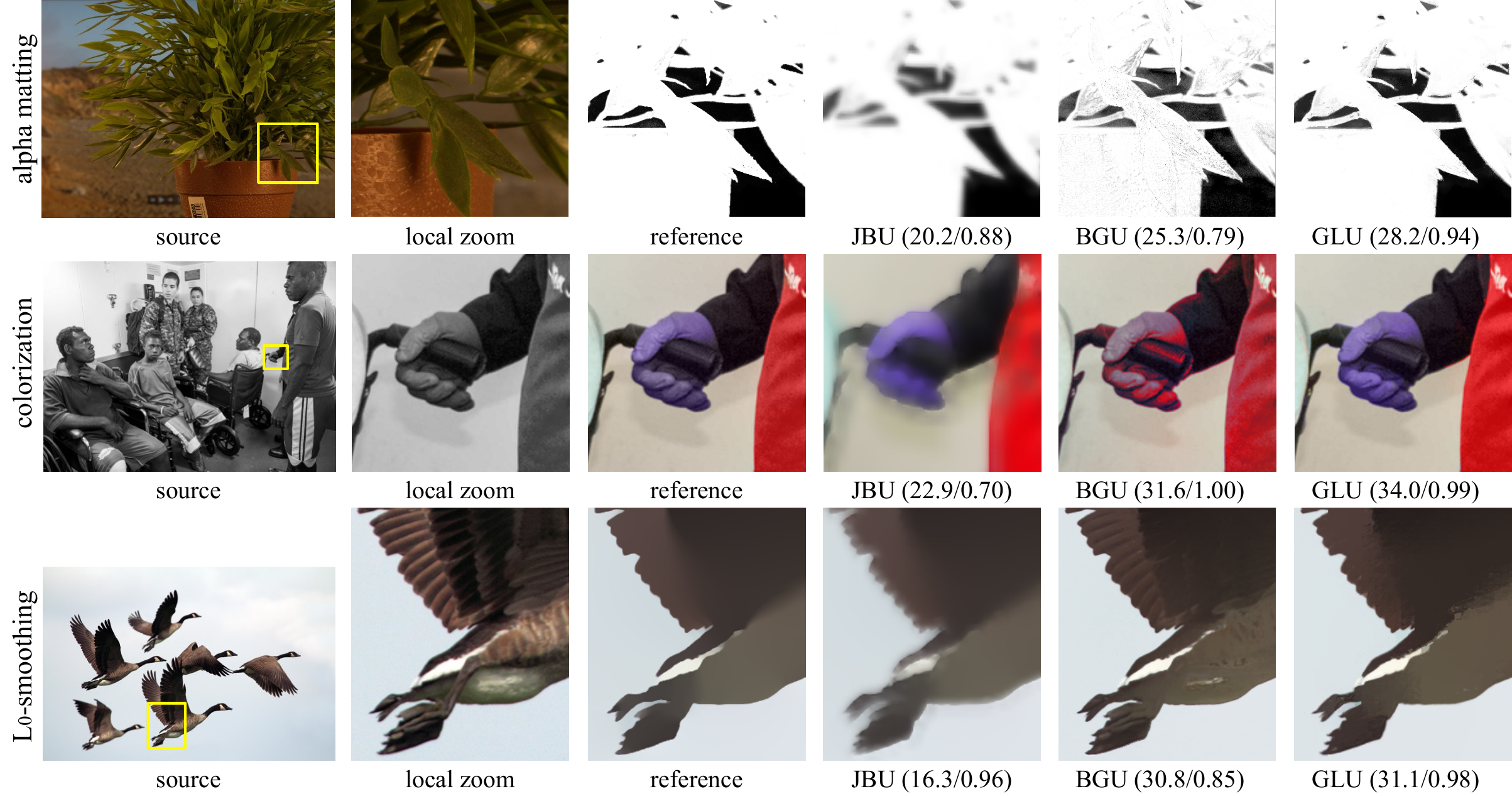}
	\caption{Visual comparisons of different methods with $8\times$ downsampling. In the parentheses are the PSNR/SSIM scores. }
	\label{fig:results}
\end{figure*}

\setlength\tabcolsep{2.0pt}
\begin{figure}[ht] 
	\centering 
        \small
        \begin{tabular}{cc}
        	\includegraphics[width=0.49\columnwidth]{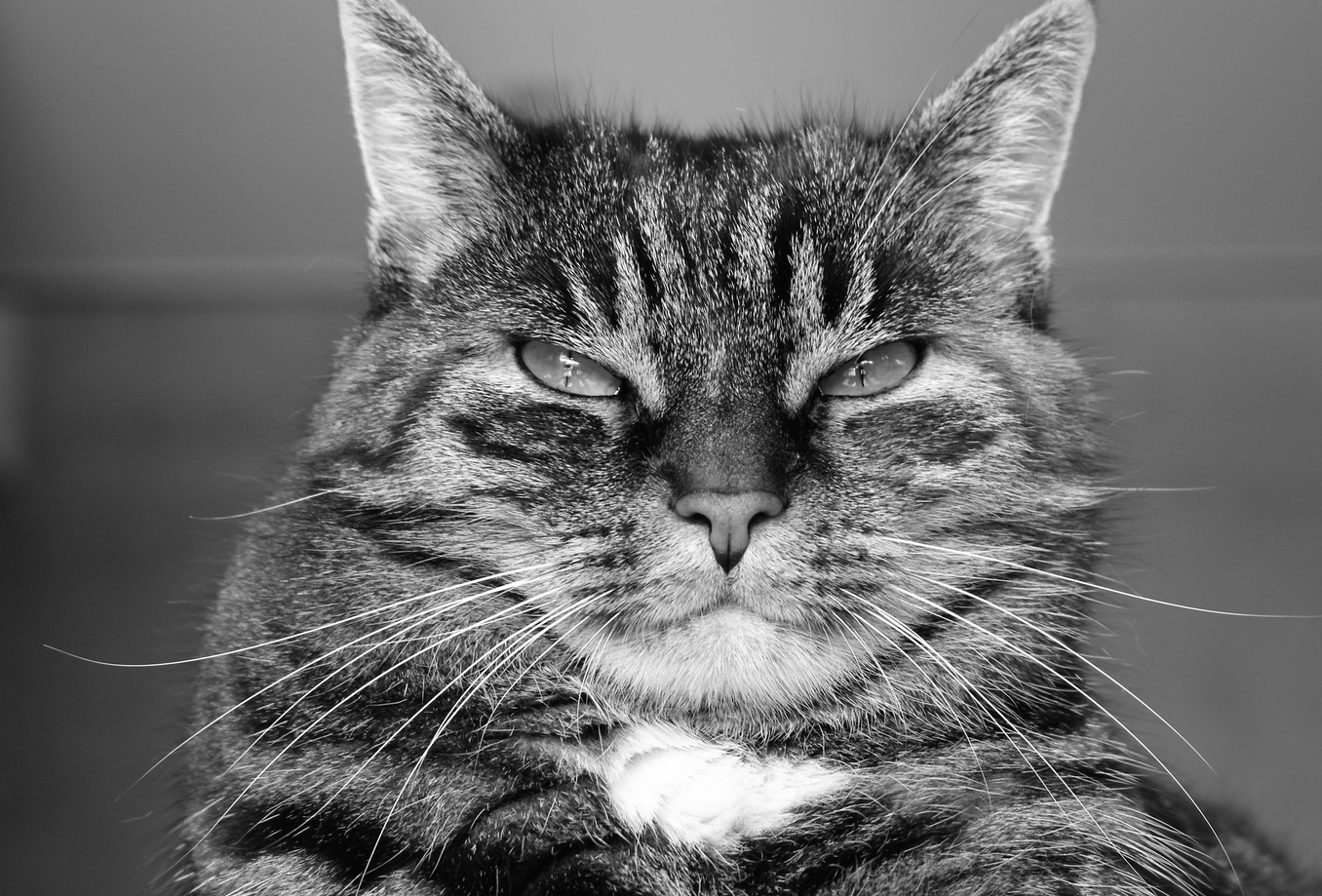} &  
        	\includegraphics[width=0.49\columnwidth]{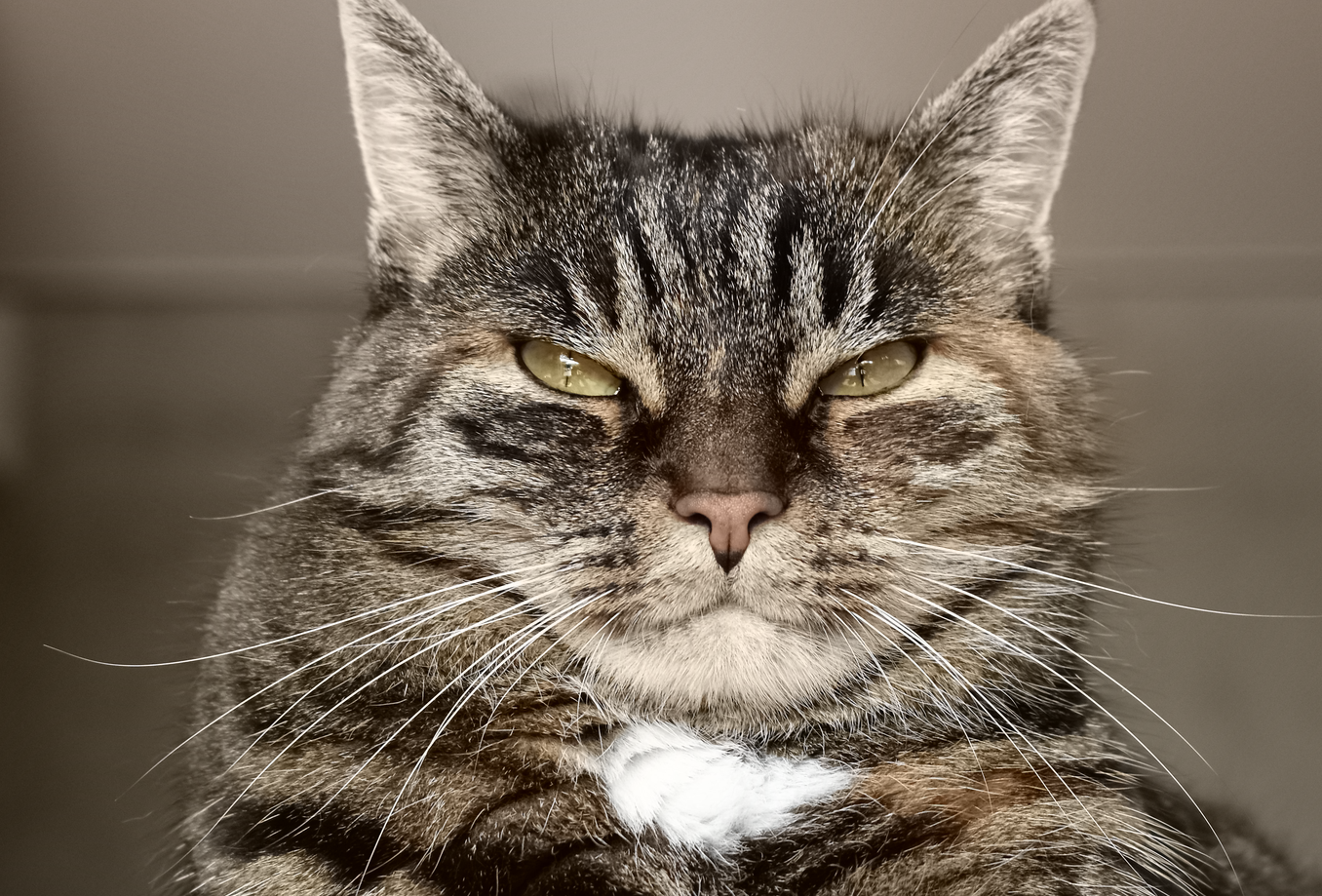} \\
        	(a)  source (1356$\times$919)   & (b) reference  \\
             \includegraphics[width=0.49\columnwidth]{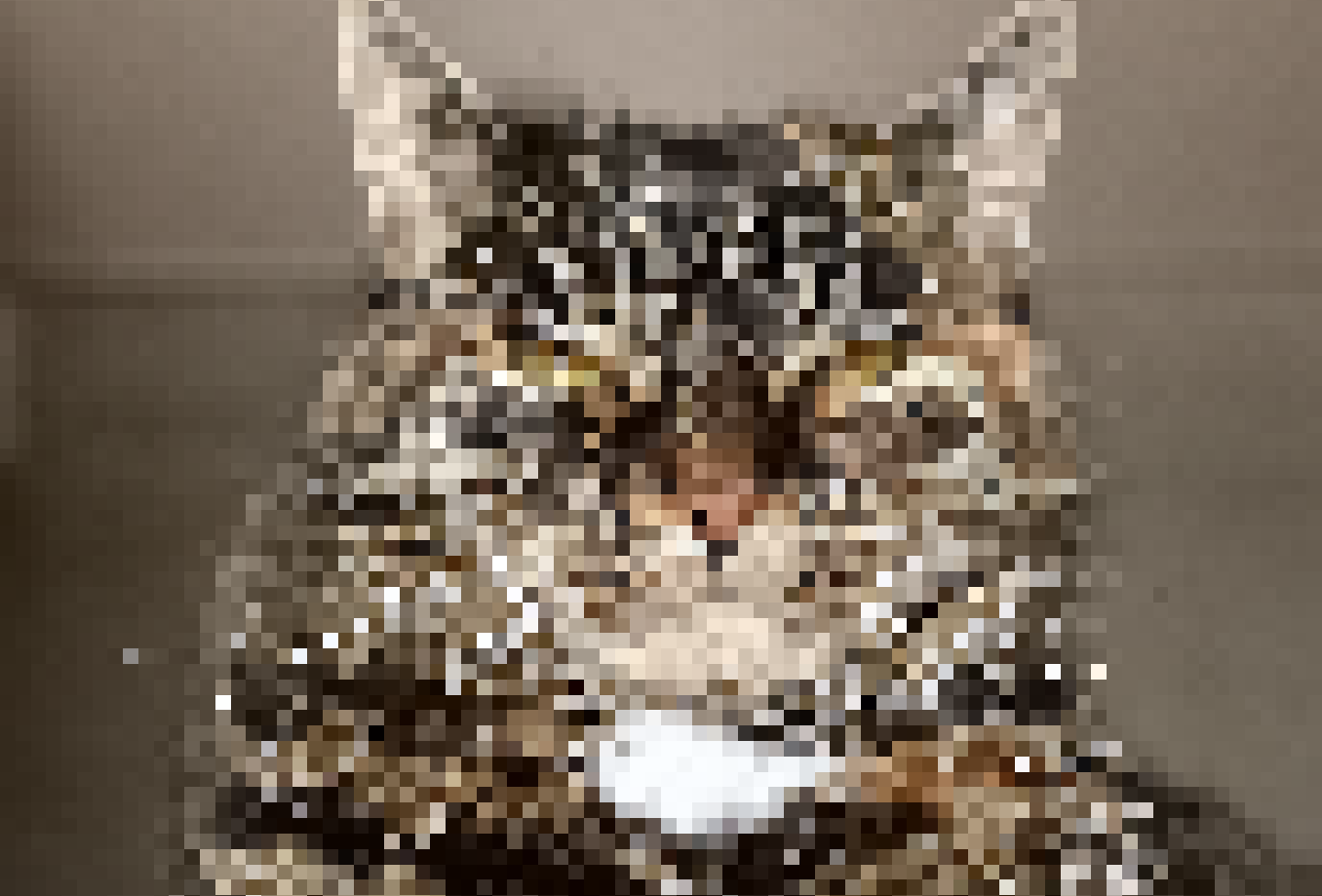} &  
             \includegraphics[width=0.49\columnwidth]{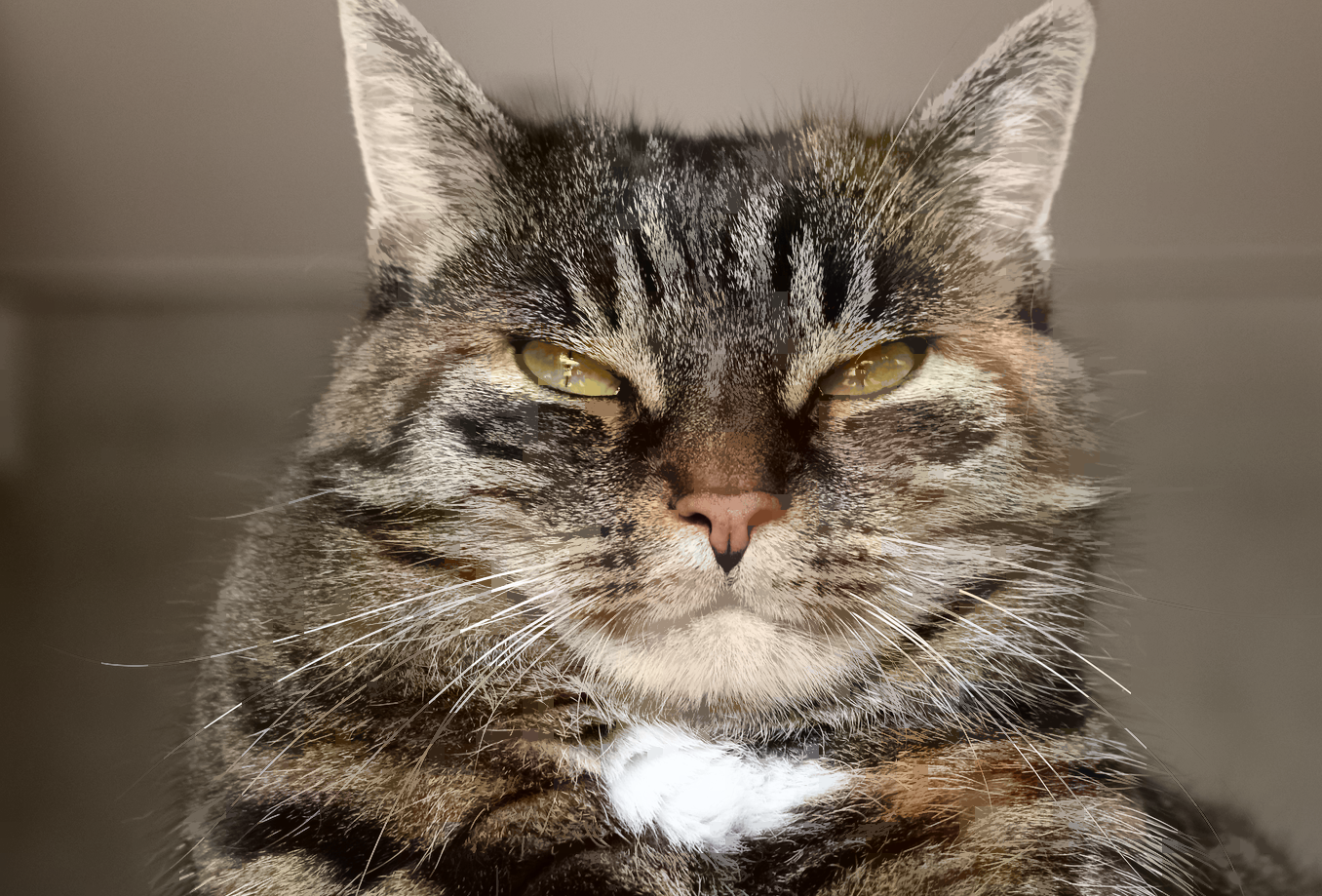} \\
             (c)  low-res target of GLU$^-$  & (d) GLU$^-$  \\
             \includegraphics[width=0.49\columnwidth]{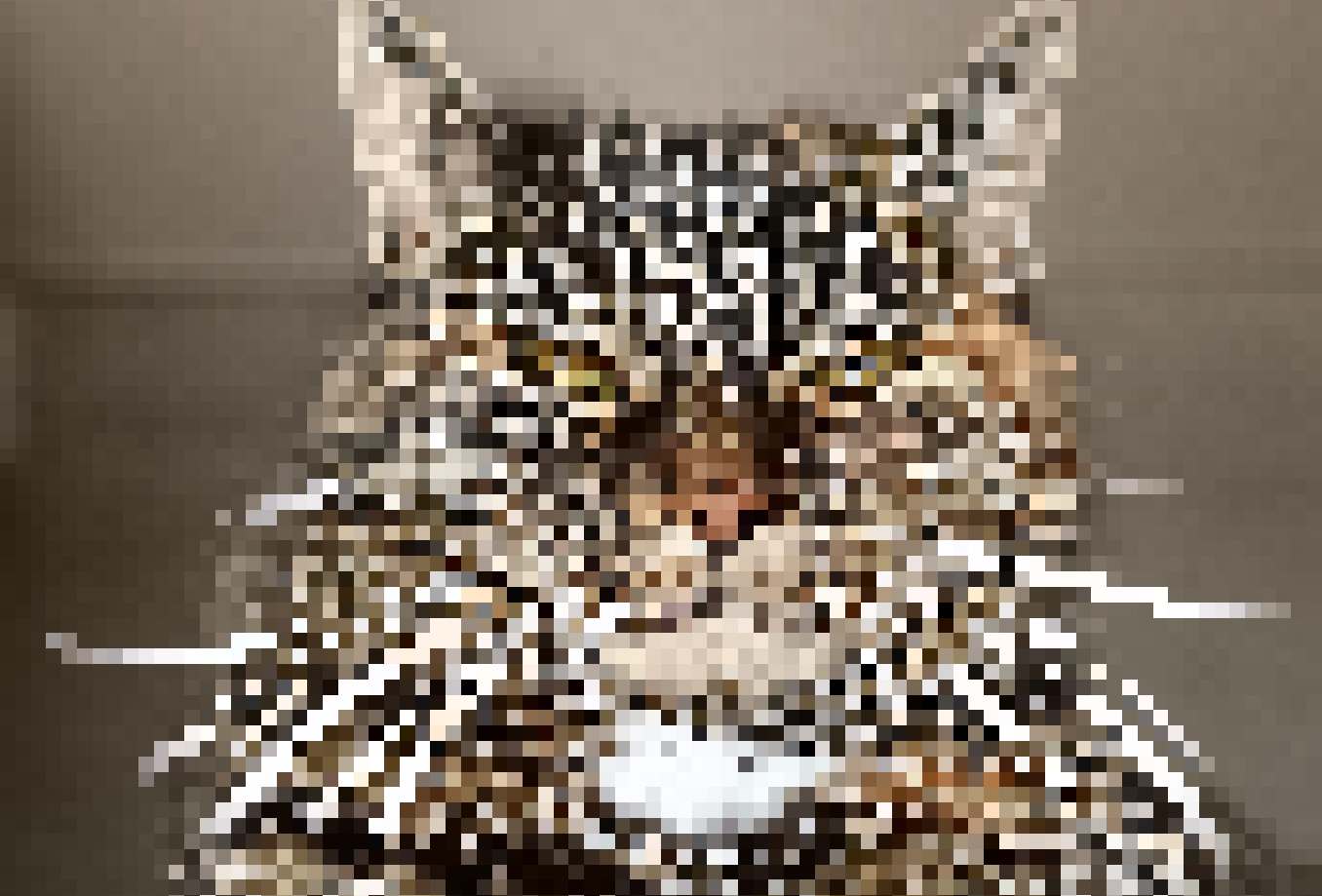} & 
             \includegraphics[width=0.49\columnwidth]{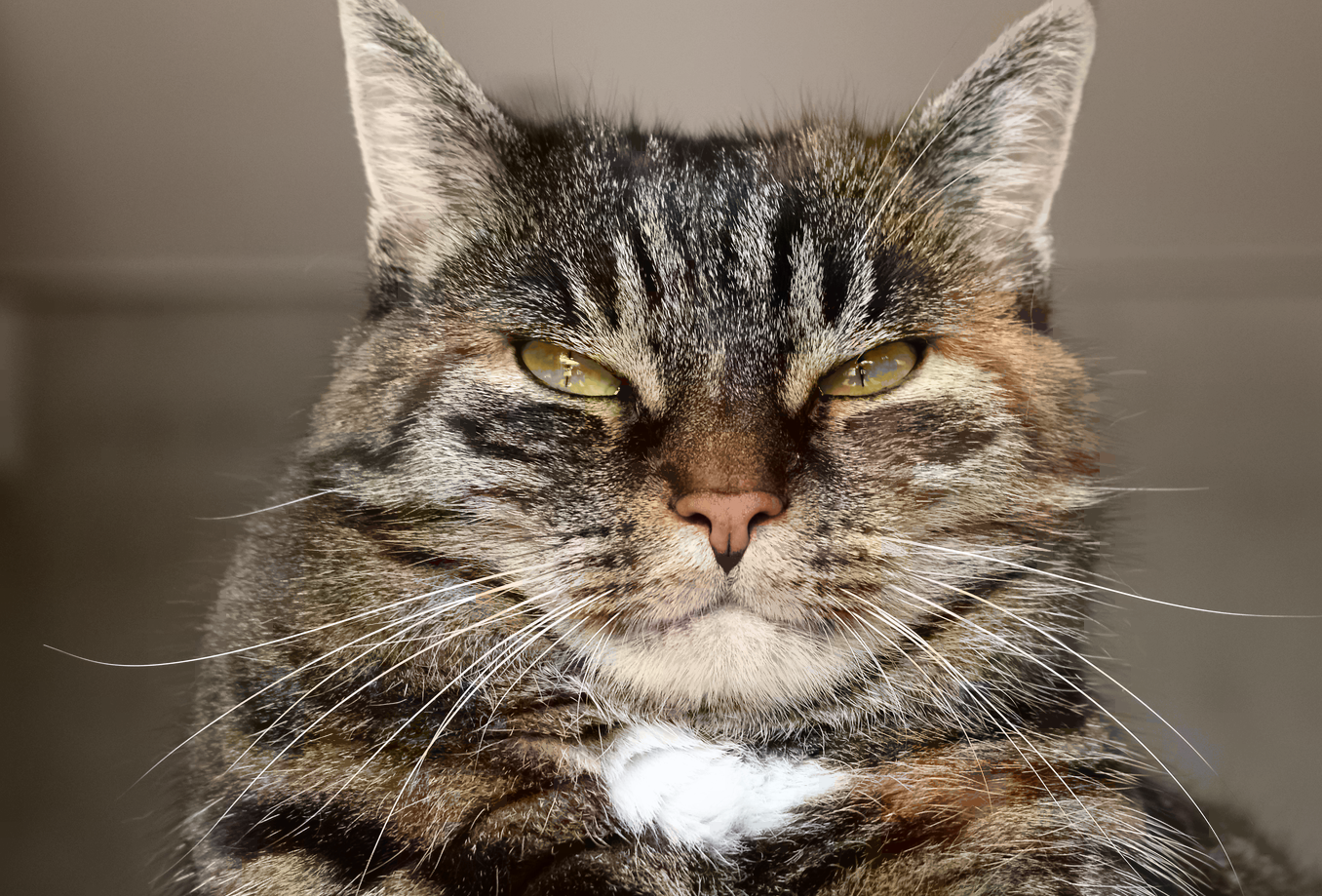} \\
             (e) low-res target of GLU  & (f) GLU  \\
        \end{tabular}
	\caption{\onerevise{The effect of downsampling optimization demonstrated on colorization at 16$\times$.} \revised{GLU can better preserve fine image structures (the cat whiskers) than GLU$^-$.}}
	\label{fig:lowres}
\end{figure}

The low-resolution target images are produced from the downsampled source images using the chosen image operators, and then upsampled \onerevise{using our technique} to produce the full-resolution target images. 
Table \ref{tab:psnr} and Table \ref{tab:ssim} are the quantitative results with PSNR and SSIM scores, respectively. PSNR measures the difference in pixel values, while SSIM mainly measures the similarity of local structures. We compare our method with JBU~\cite{kopf2007joint} and BGU~\cite{chen2016bilateral}. For JBU, we use the default parameter setting with $5\times5$ support windows and $\sigma_d=0.5, \sigma_r=0.1$. For BGU we test both the global method with the authors' MATLAB code, and the fast local method (BGU-fast) implemented with Halide~\cite{ragan2012decoupling,ragan2013halide}. For GLU we use the default setting as described in Section \ref{sec:parameters}.  GLU$^-$ is our method without downsampling optimization.  

For most applications our method outperforms JBU and BGU for both scores. \reffig{fig:results} shows some examples. In general, for large scaling ratios such as $8\times$ and $16\times$, JBU tends to overblur low-contrast edges, while BGU tends to produce bleeding artifacts that mix the effects of different regions 
due to the smooth constraint of local transformations, as we analyzed in Sections \ref{sec:edge} and \ref{sec:details}.  

Since BGU represents the target image as the local transformations of the source image, it is good at preserving the source image structures, and is therefore advantageous for applications such as \emph{Colorization}, \emph{Unsharp Masking}, and \emph{Laplacian Filter}. For these applications BGU achieves better SSIM scores than our method. In particular, for \emph{Colorization}, BGU obtains a full SSIM score, because our colorization method modifies only the chrominance channels, while SSIM is computed using only the grayscale channel. However, the PSNR scores of these applications are still comparable to or lower than our method.
For applications where some source details need to be removed, such as \emph{Matting} and \emph{Smoothing},  preserving the local structures of the source image may have unwanted effects, e.g. for image smoothing BGU may re-introduce some source details that have been removed by the smoothing operator, as demonstrated in \reffig{fig:results}. For these applications our method can significantly outperform BGU.

\revised{The downsample optimization can improve the PSNR and SSIM scores for all tested applications. As shown in \reffig{fig:lowres}, fine image structures can be better preserved with the optimized downsampling. In comparison with the regular grid downsampler, the optimized downsampler may result in more gritty and aliased low-resolution images due to the irregular spatial sampling. However, note that such aliasing actually can improve the upsampled image because the joint optimization only accepts results with \onerevise{lower} upsampling errors. Moreover, for operators that can preserve local affinities (as assumed by our method), the low-resolution target image should have the same aliasing as the downsampled source image, which should be beneficial for the upsampled target image according to our analysis in \onerevise{Section} \ref{sec:analysis2}.}

\begin{figure}
	\centering 
        \small
        \begin{tabular}{cc}
             \includegraphics[width=0.49\columnwidth]{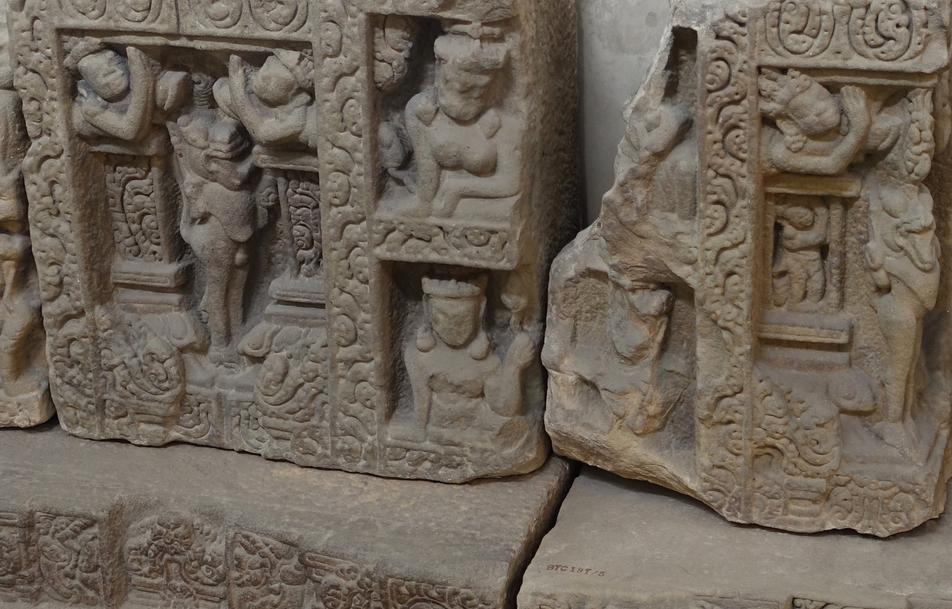} &  
             \includegraphics[width=0.49\columnwidth]{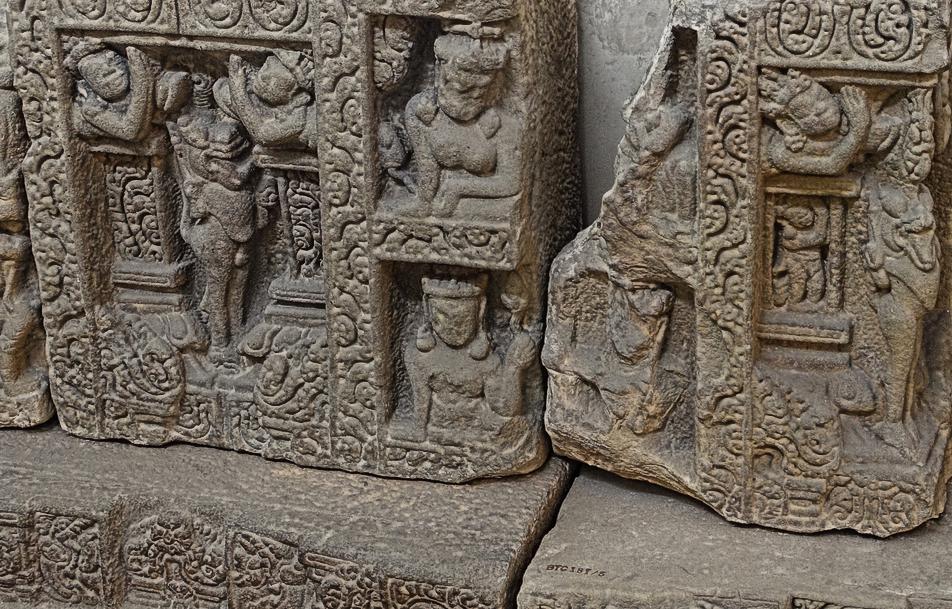} \\
             (a) source  & (b) reference  \\
             \includegraphics[width=0.49\columnwidth]{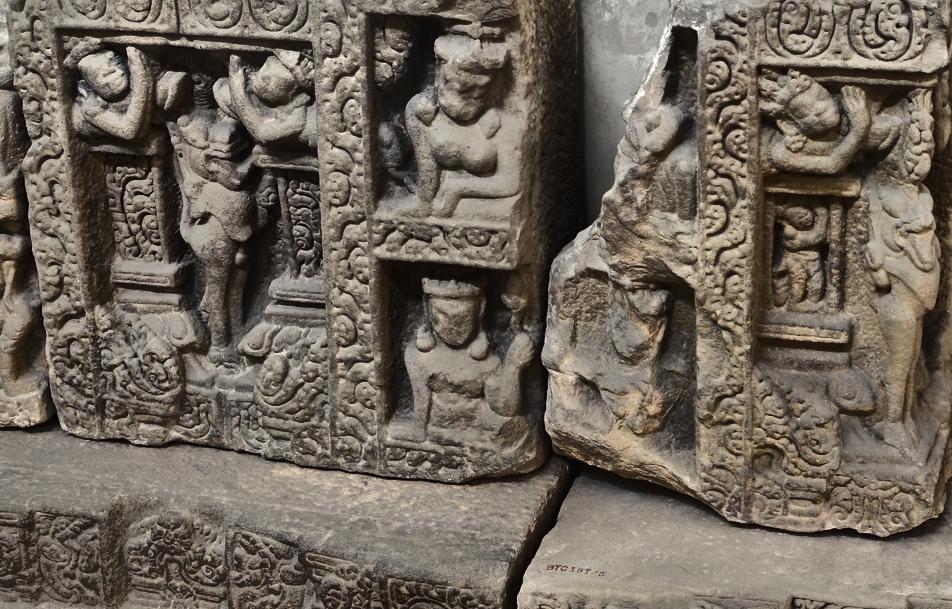} & 
             \includegraphics[width=0.49\columnwidth]{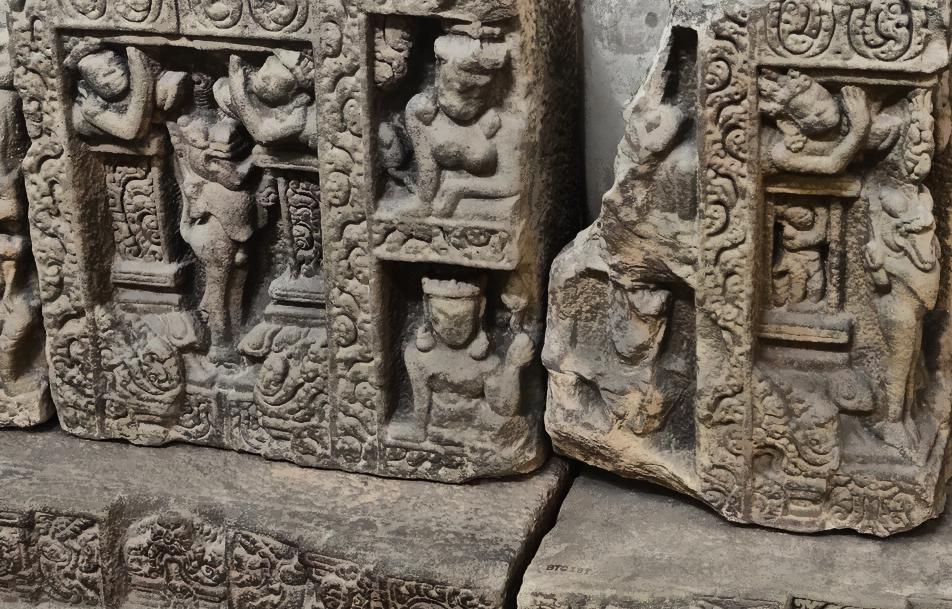} \\
             (c) BGU & (d) GLU \\
        \end{tabular}
        %\vspace{-3mm}
	\caption{\emph{Unsharp Masking} with $8\times$ downsampling. The results of BGU and GLU are comparable, and both look better than the reference.}
	\label{fig:app-um}
\end{figure}

For \emph{Unsharp Masking}, it is strange that BGU-fast performs better than BGU in terms of PSNR. Actually, this is caused by the different effects of the image operator at different scales. As shown in \reffig{fig:app-um}, the upsampled results of BGU and GLU are largely different from the reference, which is obviously caused by the image operator rather than the upsampler. In this case, \onerevise{we consider the scores to be mostly noise} and not trustworthy for the evaluation. The last column of \onerevise{Tables} \ref{tab:psnr} and \ref{tab:ssim} shows the results with low-resolution target downsampled from the reference, which can better reflect the net effect of upsampling methods.

\begin{figure*}
	\centering
	\small
	\begin{tabular}{cccc}		
		\includegraphics[width=0.23\linewidth]{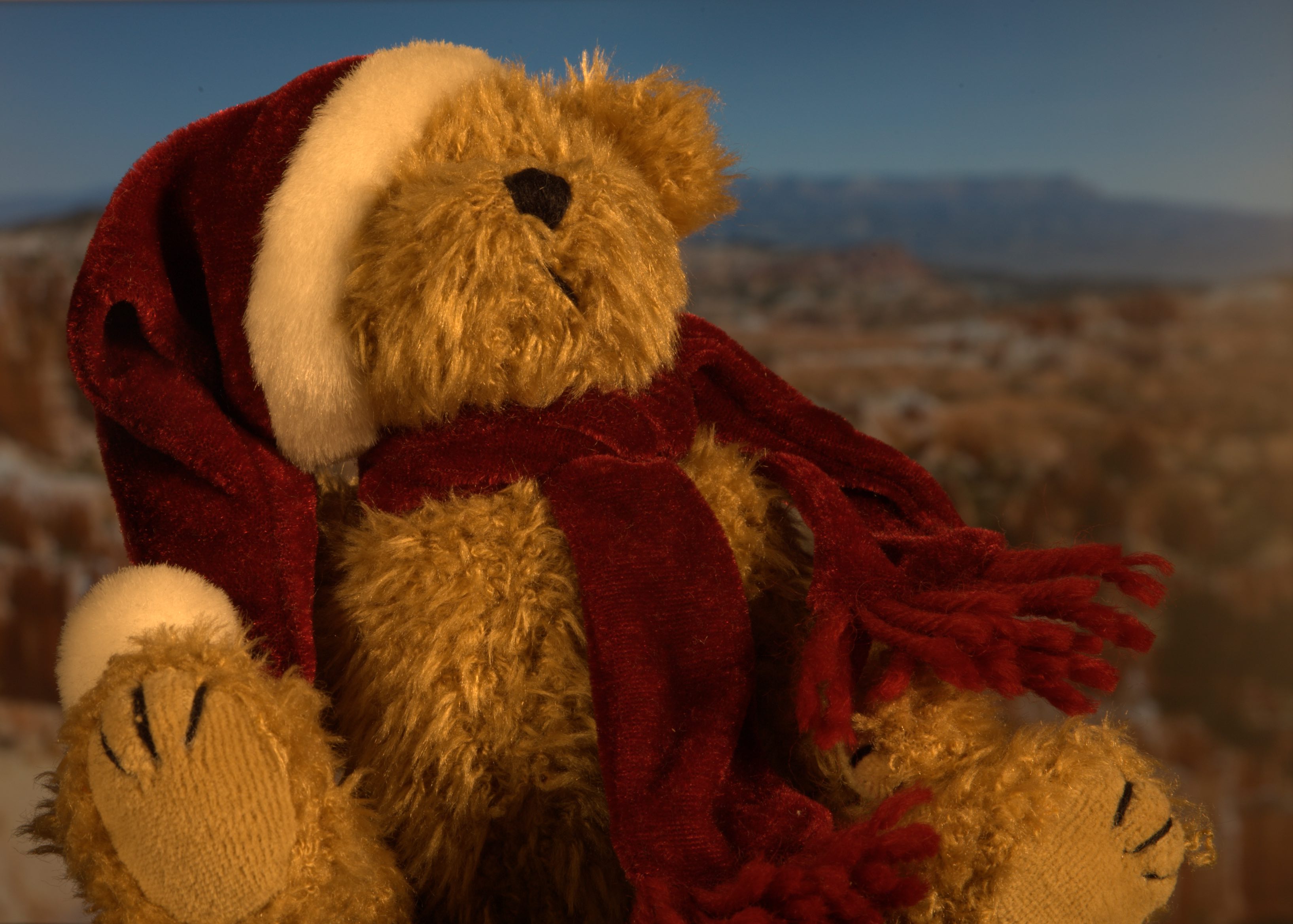} & 
		\includegraphics[width=0.23\linewidth]{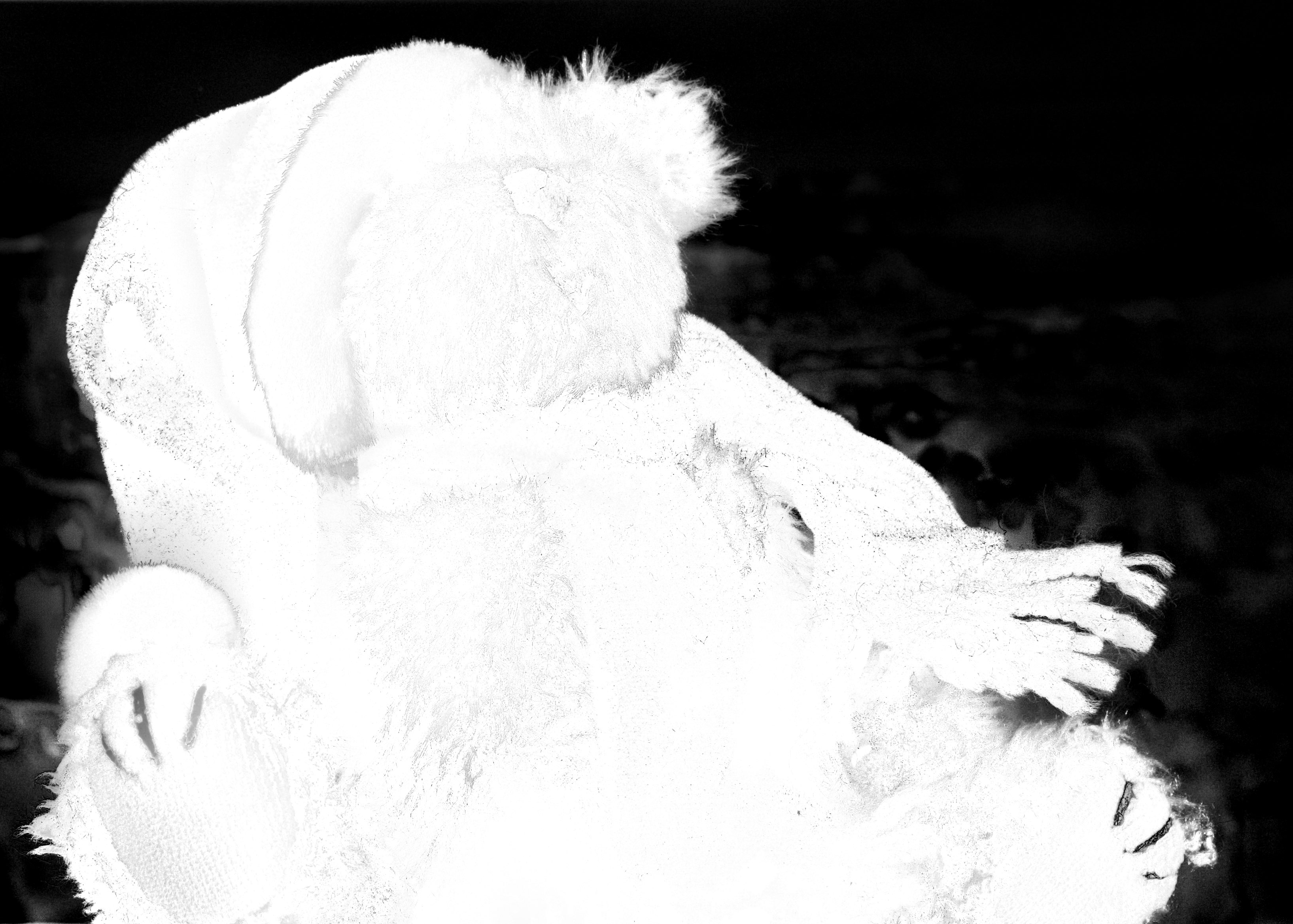} &
		\includegraphics[width=0.23\linewidth]{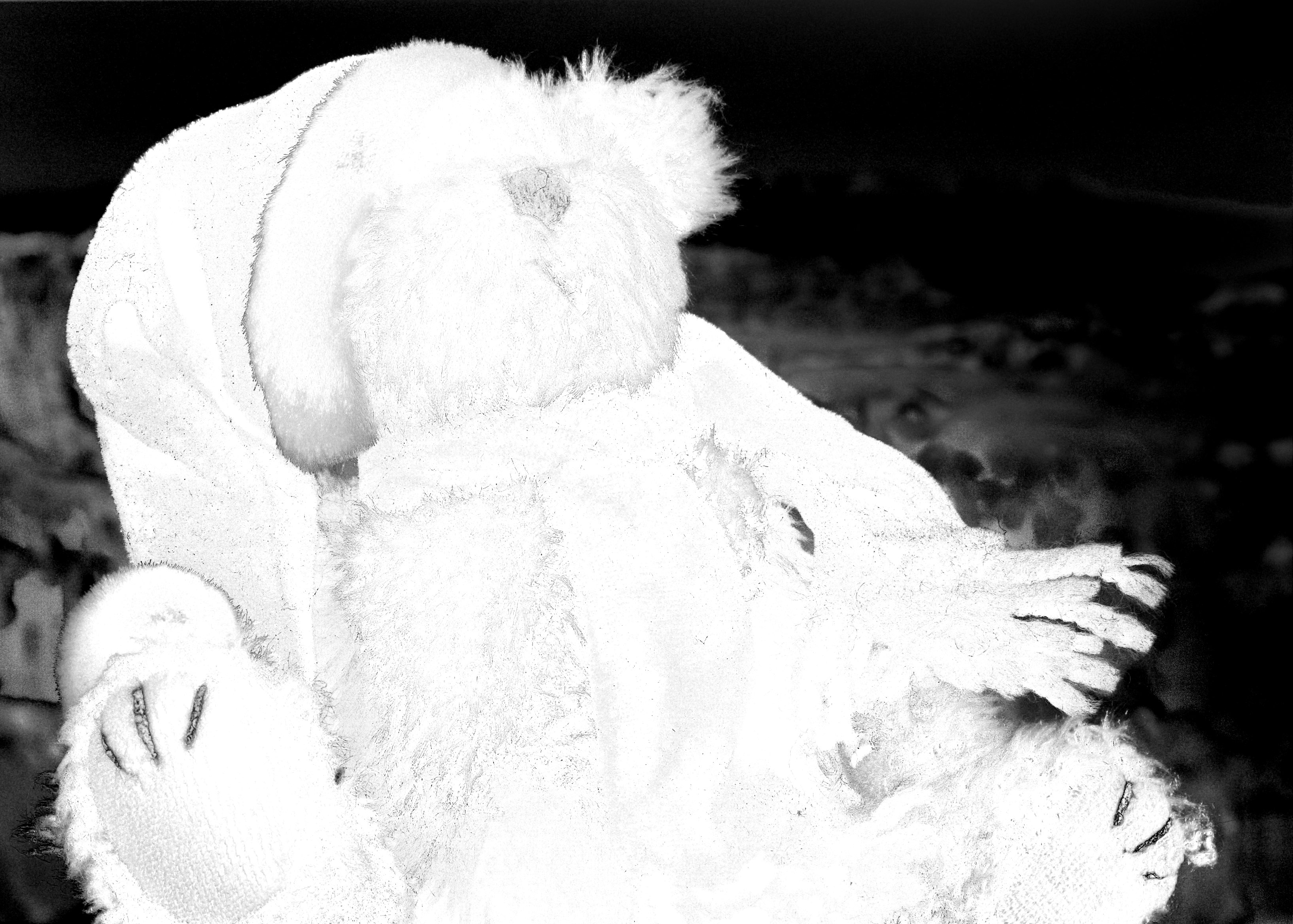} &
		\includegraphics[width=0.23\linewidth]{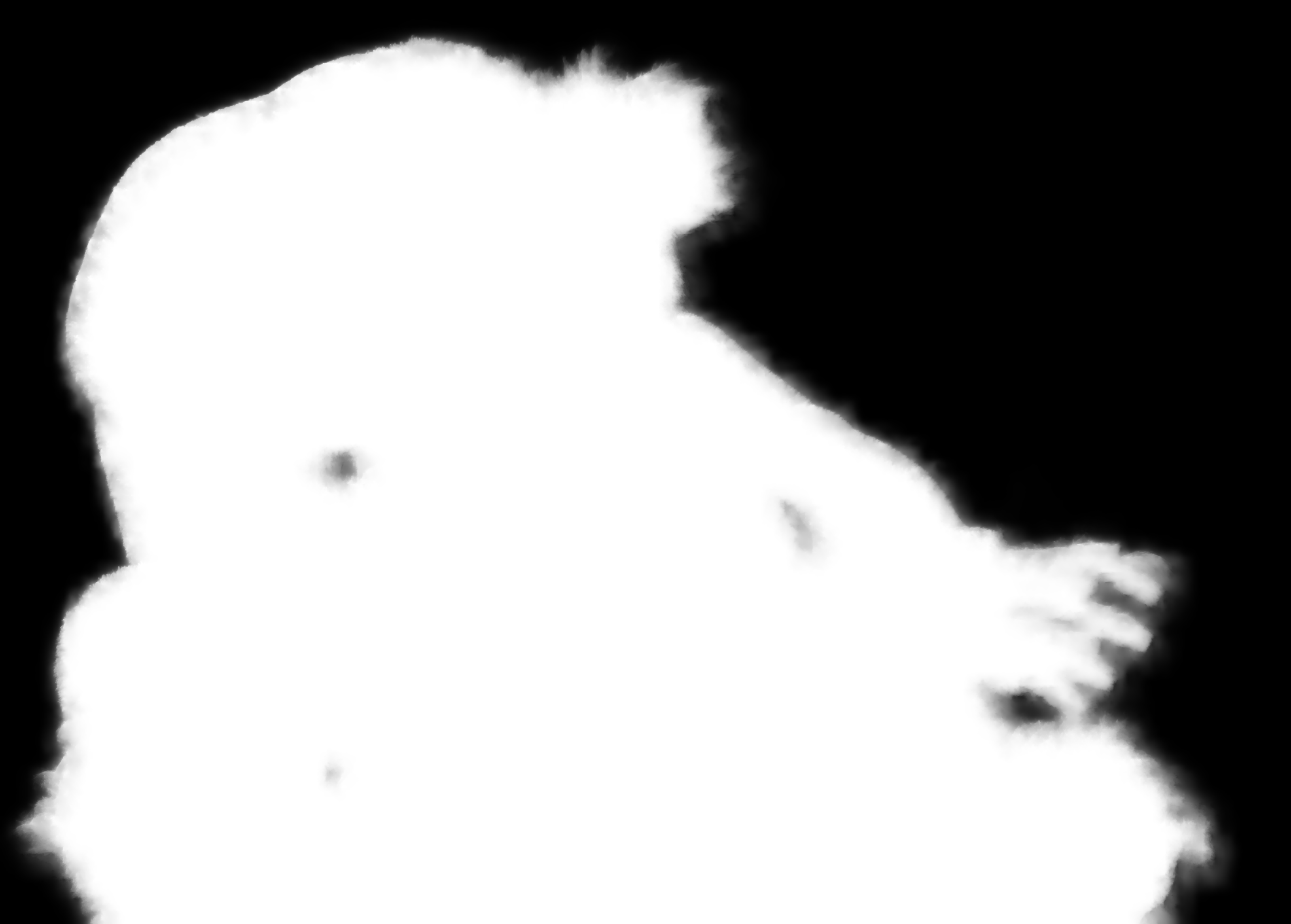}  \\
		(a) Source ($3270\times 2388$) & (b) BGU 32$\times$ & (c) BGU 64$\times$ & (d) JBU 32$\times$ \\
		
		\includegraphics[width=0.23\linewidth]{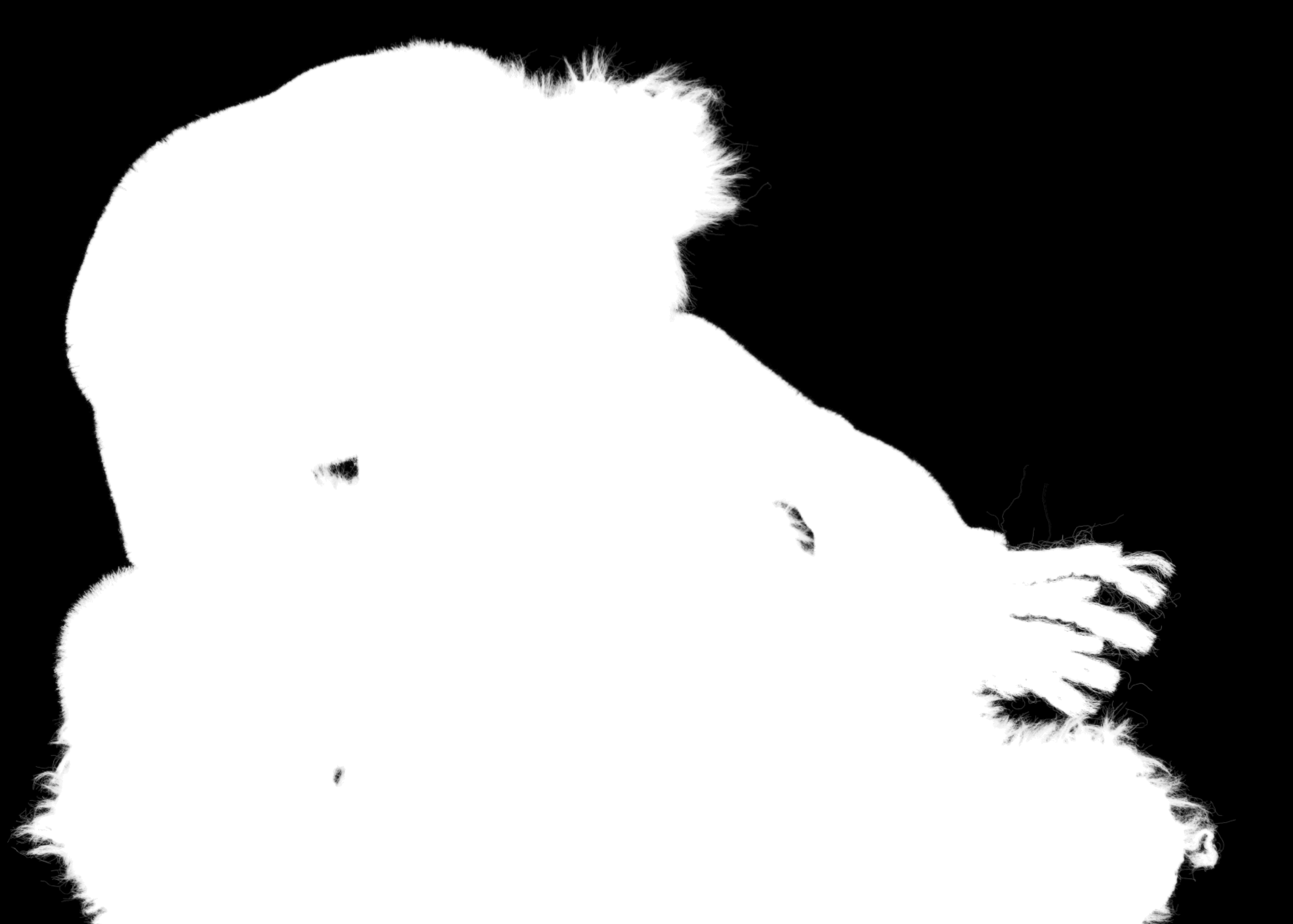} & 
		\includegraphics[width=0.23\linewidth]{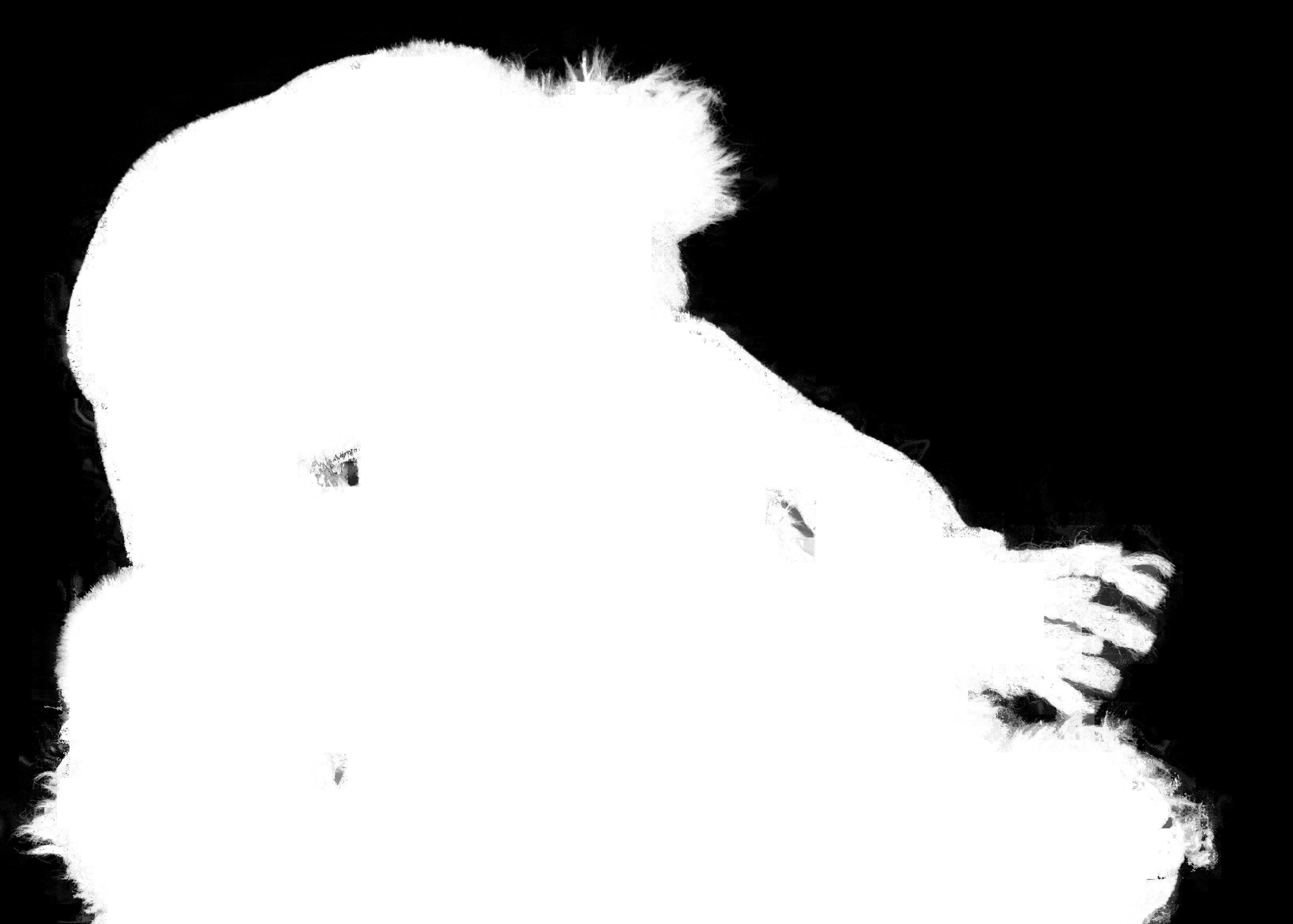} &
		\includegraphics[width=0.23\linewidth]{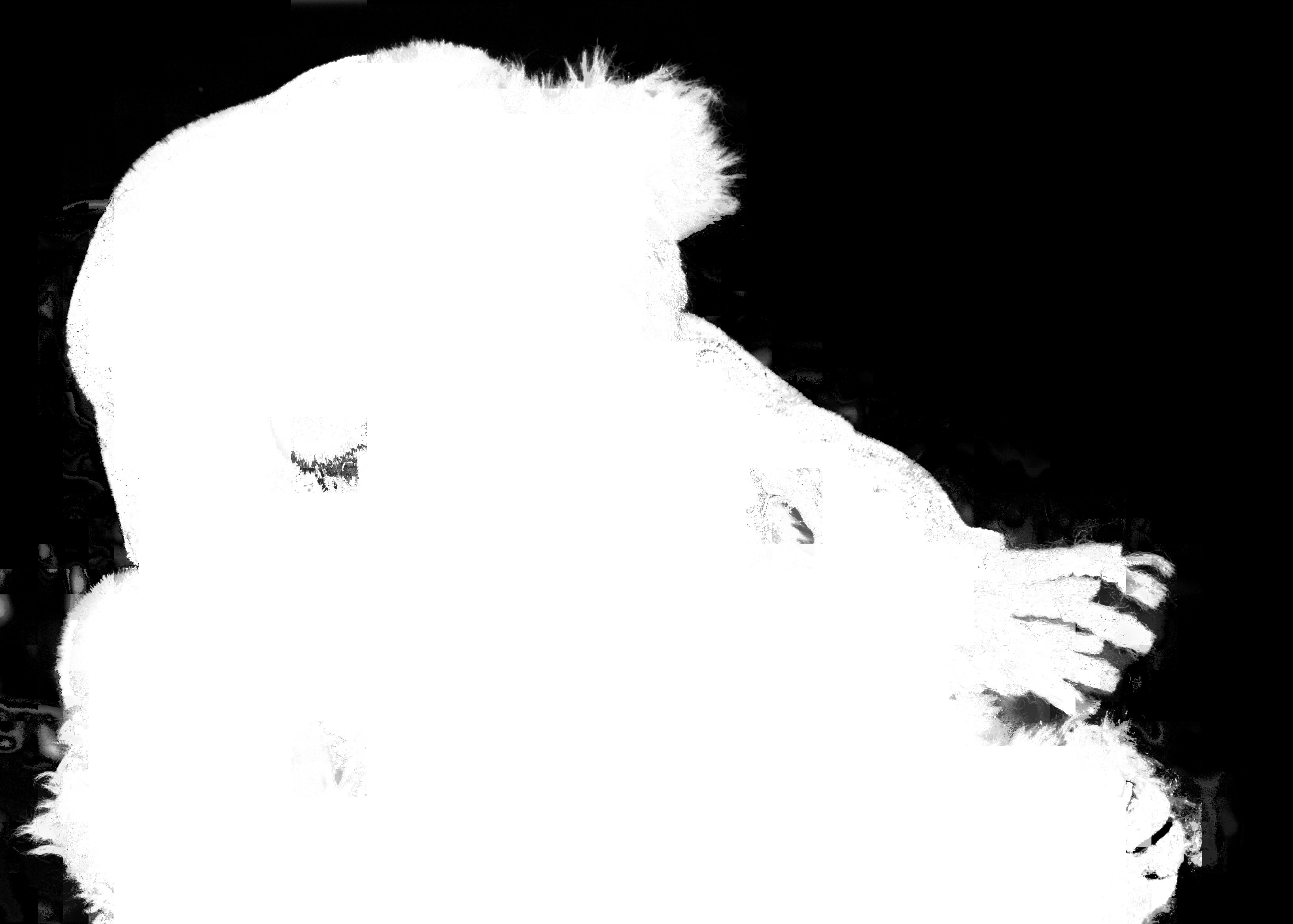} &
		\includegraphics[width=0.23\linewidth]{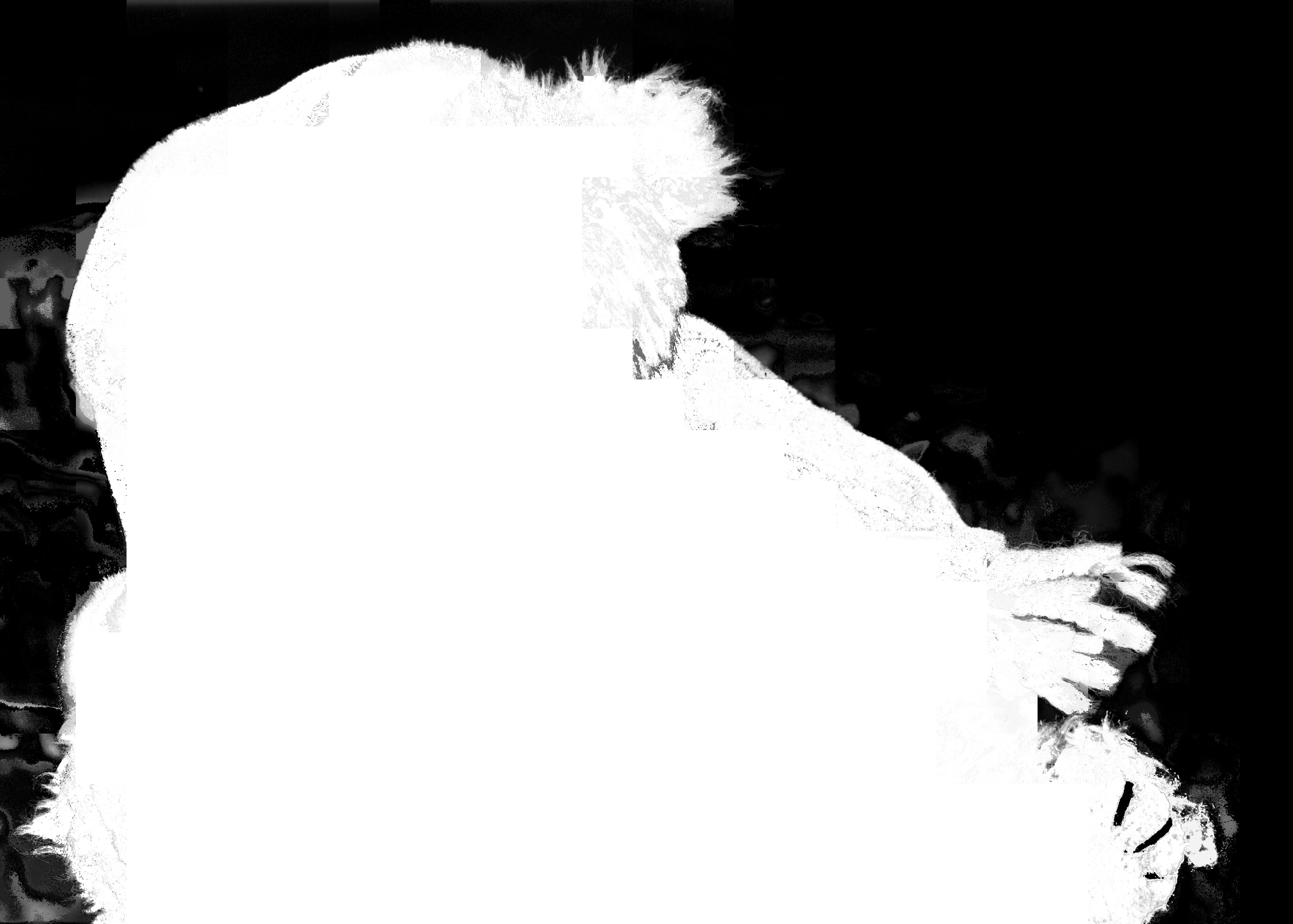} \\
		
		(e) Reference & (f) GLU 32$\times$ & (g) GLU 64$\times$ & (h) GLU 128$\times$\\
	\end{tabular}
        %\vspace{-3mm}
	\caption{Experiments with large ratios of downsampling and upsampling. JBU produces obvious blur for $32\times$, BGU produces significant bleeding for $64\times$, and our method gets pretty good results even for $128\times$. For this experiment the low-resolution target images are downsampled from the reference, in order to observe the net effect of upsampling methods. }
	\label{fig:nice}
\end{figure*}

Finally, to see the effect of our method for large ratio of upsampling, we conducted experiments with larger ratios of $32\times$, $64\times$, and even $128\times$. \reffig{fig:nice} shows an example for matting. Surprisingly, for the tested image, even for $128\times$ of downsampling and upsampling, our method still obtains pretty good results, which are better than the results of JBU and BGU with smaller ratios.

\setlength\tabcolsep{5.0pt}

\subsection{Parameters}
\label{sec:parameters}

%\subsubsection{Performance and Accuracy}
%Shown as Table ~\ref{tab:psnr} and Table ~\ref{tab:ssim}, we compare our upsampling proposal with JBU and BGU on multiple datasets for natural image editing operators: matting, denoising et. al. Results in Table ~\ref{tab:psnr} show that our algorithm performs better on upsampling than other algorithms except for recolor (although BGU has the highest quantitative result in recoloring, qualitative results show that it contains coloring that is inconsistent with the target image). And results in Table ~\ref{tab:ssim} show that the structure-preserving ability of our method is as good as that of JBU and BGU, which indicates that our method is meaningful as upsampling.

Our method contains only 3 parameters, i.e. the window size $S$ of $\Omegaplow$, the error threshold $\tau$ and the maximum number of iterations $N$. $\tau$ should be an error value that may introduce noticeable visual difference, and for most examples our joint optimization requires only 1 or 2 iterations to remove large-error regions, so the parameters are easy to set. In our experiments, they are fixed to $S=3$, $\tau=30/255$, $N=3$. 

\renewcommand{\arraystretch}{1.2} 
\begin{table}
  \caption{\modified{Effect of the window size $S$ to the resulting quality in PSNR.}}
  \vspace{-3mm}
  \label{tab:patchsize}
  %\resizebox{0.7\columnwidth}{!}{
  \begin{tabular}{lllll}
    \toprule
    window size&$3\times3$&$5\times5$&$7\times7$ & $9\times9$\\
    \midrule
    self-upsampling & 41.31 & 42.83 & 44.04 & \textbf{44.92}\\
    matting & \textbf{35.47} & 35.18 & 34.76 & 34.15\\
    colorization & \textbf{32.39} & 29.95 & 29.94 & 28.74\\
  \bottomrule
\end{tabular}%}
\end{table}

Table \ref{tab:patchsize} investigates the effect of the window size $S$. Using a larger window size can lead to better self-upsampling, but slightly reduces the quality of the target images. Actually, a $3\times 3$ window in the downscaled image corresponds to a large enough neighborhood in the original resolution, so using a larger window size is not necessary and may degrade performance around low-contrast edges.

Compared with JBU and BGU, the parameter setting of our method is much simpler. This advantage makes it more suitable to be used as a universal guided upsampler for different situations.

\begin{table}
    \caption{\modified{Time cost (ms).}}
    \vspace{-3mm}
    \begin{tabular}{cccccc}
    \toprule
    \multirow{2}*{Image Size} & JBU
		& BGU &  BGU-fast & GLU & GLU$^-$   \\ 
		% \cmidrule(r){2}\cmidrule(r){3}\cmidrule(r){4}\cmidrule(r){5}\cmidrule(r){6}
		& \small{C++} &   \small{Matlab}  &  \small{Halide} &   \small{C++} &   \small{CUDA} \\
    \midrule
        2K  & 364	& 5772  & 13.1  &	125 & \textbf{5.6}\\
        4K  & 1427 &	18029  & 28.5 &  507	& \textbf{14.3}\\ 
    \bottomrule
    \end{tabular}
     \label{tab:timecost}
\end{table}

\renewcommand{\arraystretch}{1.0} 

\subsection{Time Cost}
\label{sec:time}

Table \ref{tab:timecost} compares the time cost of different methods. The BGU global method implemented with \onerevise{MATLAB} is slow to compute, JBU and GLU are much faster, but still cannot reach real-time speed. BGU-fast implemented with Halide can be very fast even on CPU, but as compared in Tables \ref{tab:psnr} and \ref{tab:ssim}, its quality is significantly lower than our method because the local method is unstable in some situations.

The main computation of our method is the joint optimization of $\Theta$ and $\Ilow$. Note that since each pixel is optimized independently, our method is parallelizable and can be greatly accelerated with GPU. For a test we implemented GLU$^-$ with CUDA and tested it on a laptop with Nvidia GTX1650 GPU. The accelerated version takes only about 5ms for 2K images and 14ms for 4K images, so it can be easily incorporated for real-time video processing.

A great advantage of our method is that the joint optimization process is \emph{target-free}, i.e. the optimization is independent of the target image. Therefore, it can be precomputed before the target image is acquired,
as we will demonstrate in Sections \ref{sec:interactive} and \ref{sec:realtime}. \revised{For applications where multiple operators may be applied for the same image, \onerevise{optimized parameters can be cached and shared between different operators}, which further reduces the overall computations of our method.}

%The downsampling optimization operates for only the regions of large error, so it would introduce only a little more computations. As mentioned above, for most image using 1 or 2 iterations is enough. Actually, after the initialization only a  small fraction of pixels need to be further processed. Given $\Theta$, the upsampling requires only a simple linear interpolation for each pixel, and the computation is negligible in most situations.
%\reffig{fig:...} demonstrate a the iteration process for an example, for which the downsampling optimization incurs only ???ms more time.

 \begin{figure}[ht]
	\centering
	\begin{tabular}{cc}
		\includegraphics[width=0.45\linewidth]{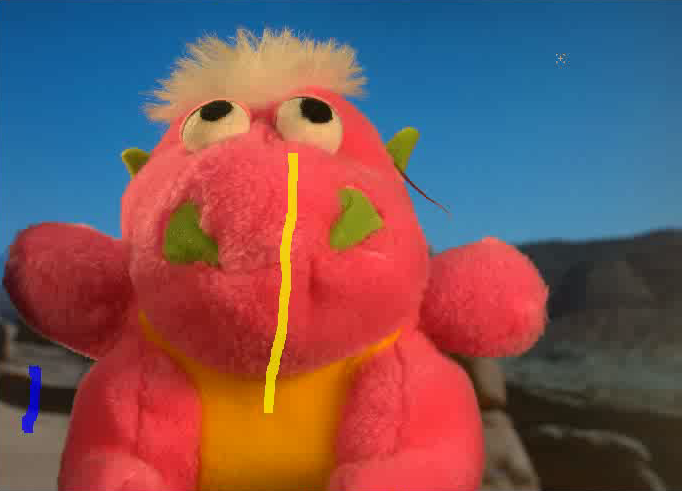} & \includegraphics[width=0.45\linewidth]{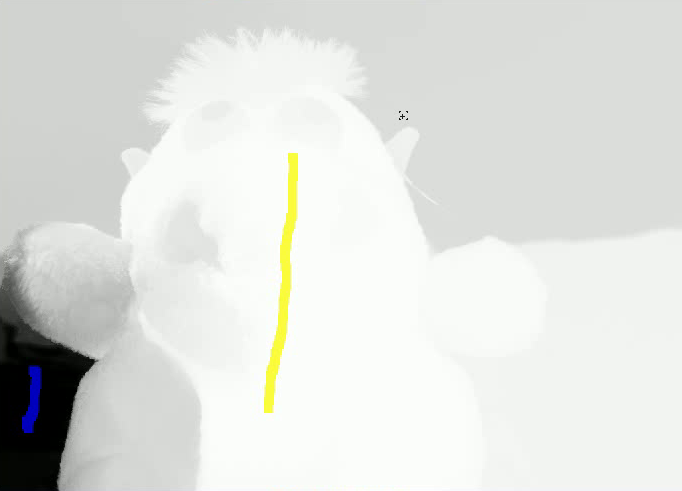} \\
		(a) & (b) \\
		\includegraphics[width=0.45\linewidth]{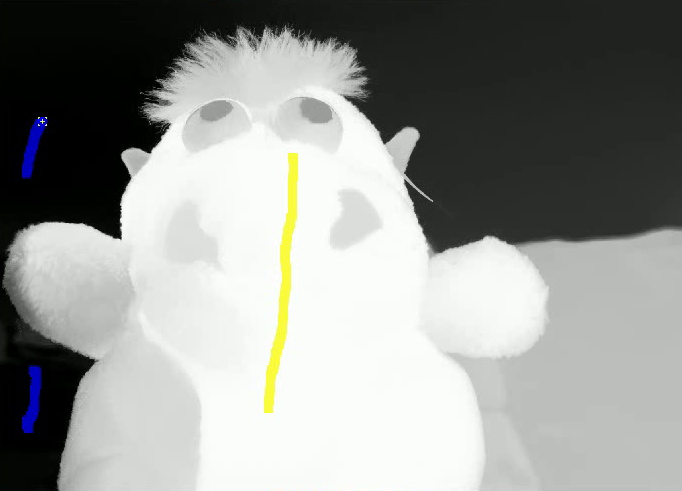} & \includegraphics[width=0.45\linewidth]{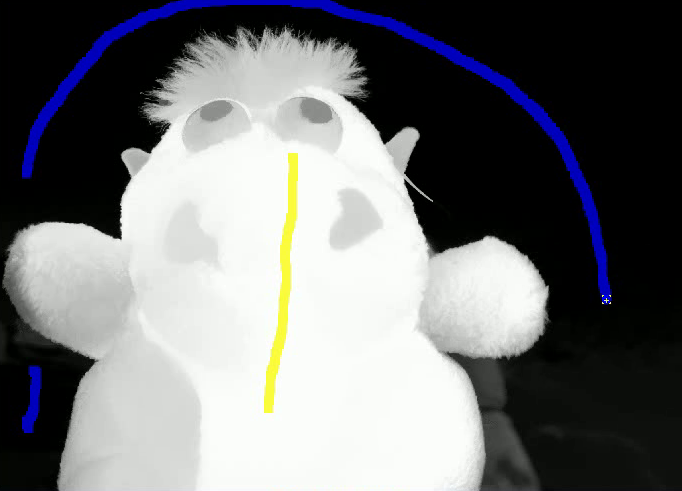}  \\
		(c) & (d)
	\end{tabular}
       \vspace{-3mm}
	\caption{Interactive matting with instant feedback. (a) input image with initial scribble brushes. (b) the initial alpha matte. (c)(d) the changes in alpha matte can be observed instantly while the user moving mouse to add a new brush. }
	\label{fig:interaction}
\end{figure}

\begin{figure}[ht]
	\centering 
	\small
	\begin{tabular}{ccc}
		\includegraphics[width=0.46\columnwidth]{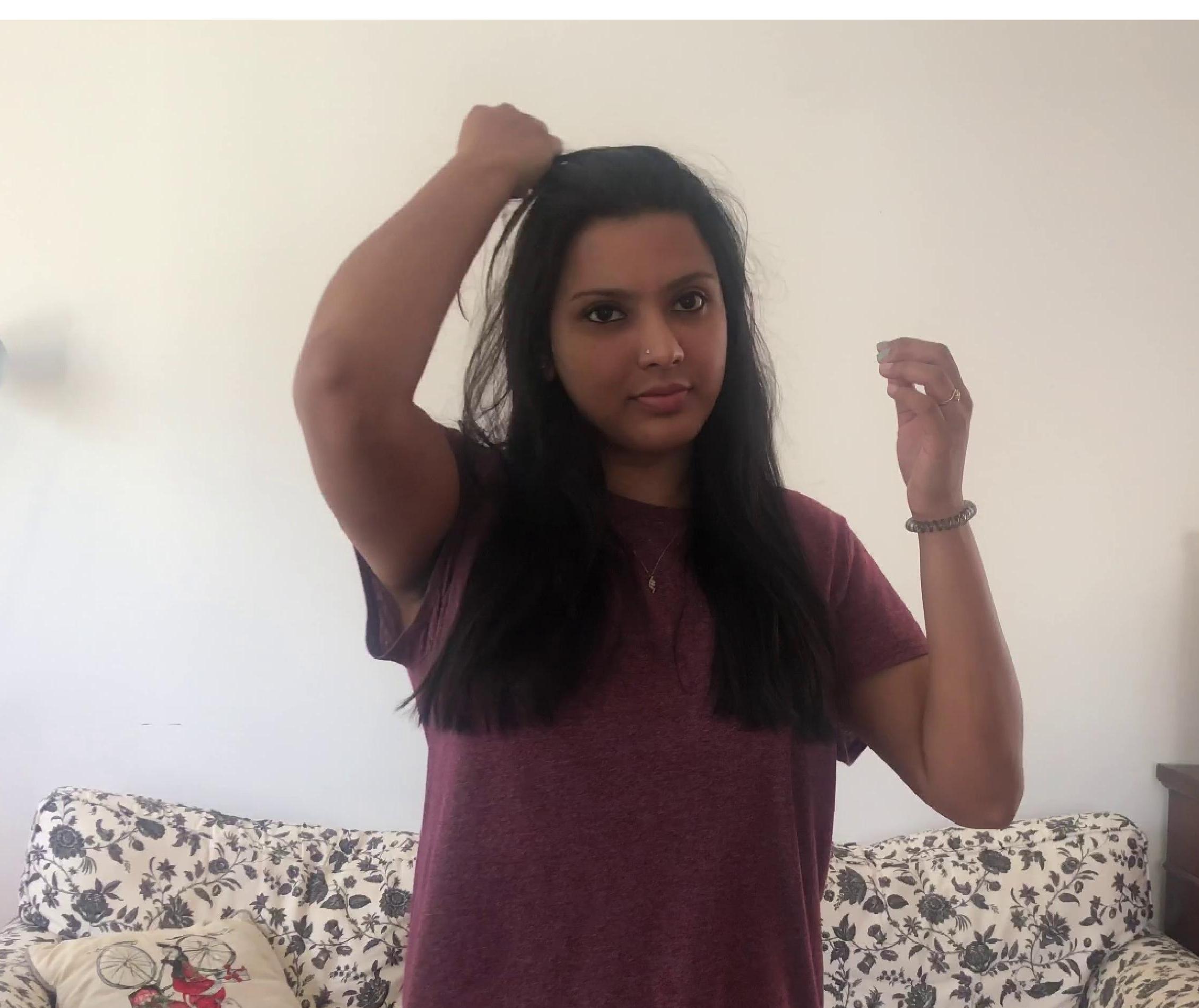} &
		\includegraphics[width=0.46\columnwidth]{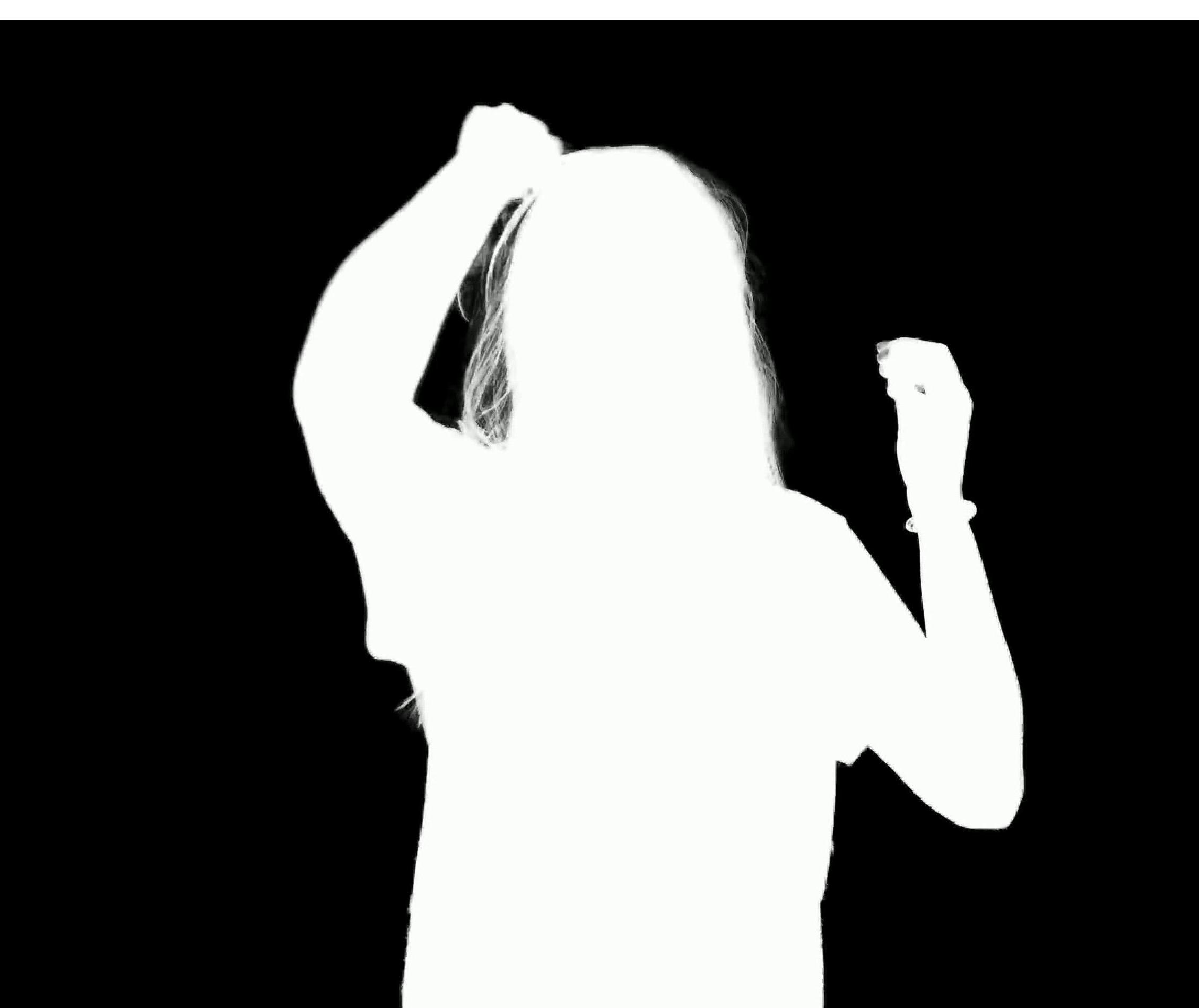} \\
		(a) Source (3840$\times$2160)  & (b) full res. (<5fps) \\ 
		\includegraphics[width=0.46\columnwidth]{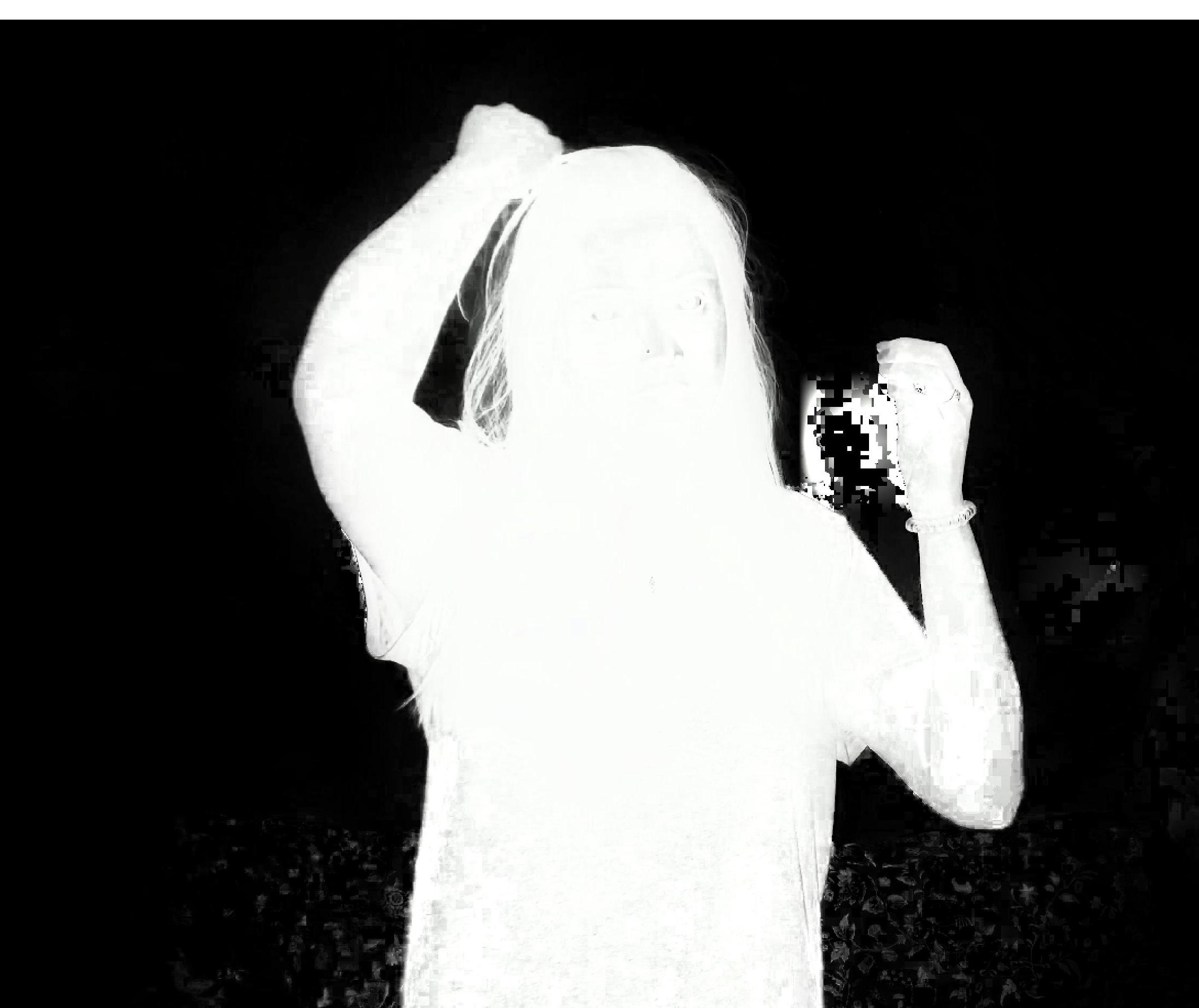} &
		\includegraphics[width=0.46\columnwidth]{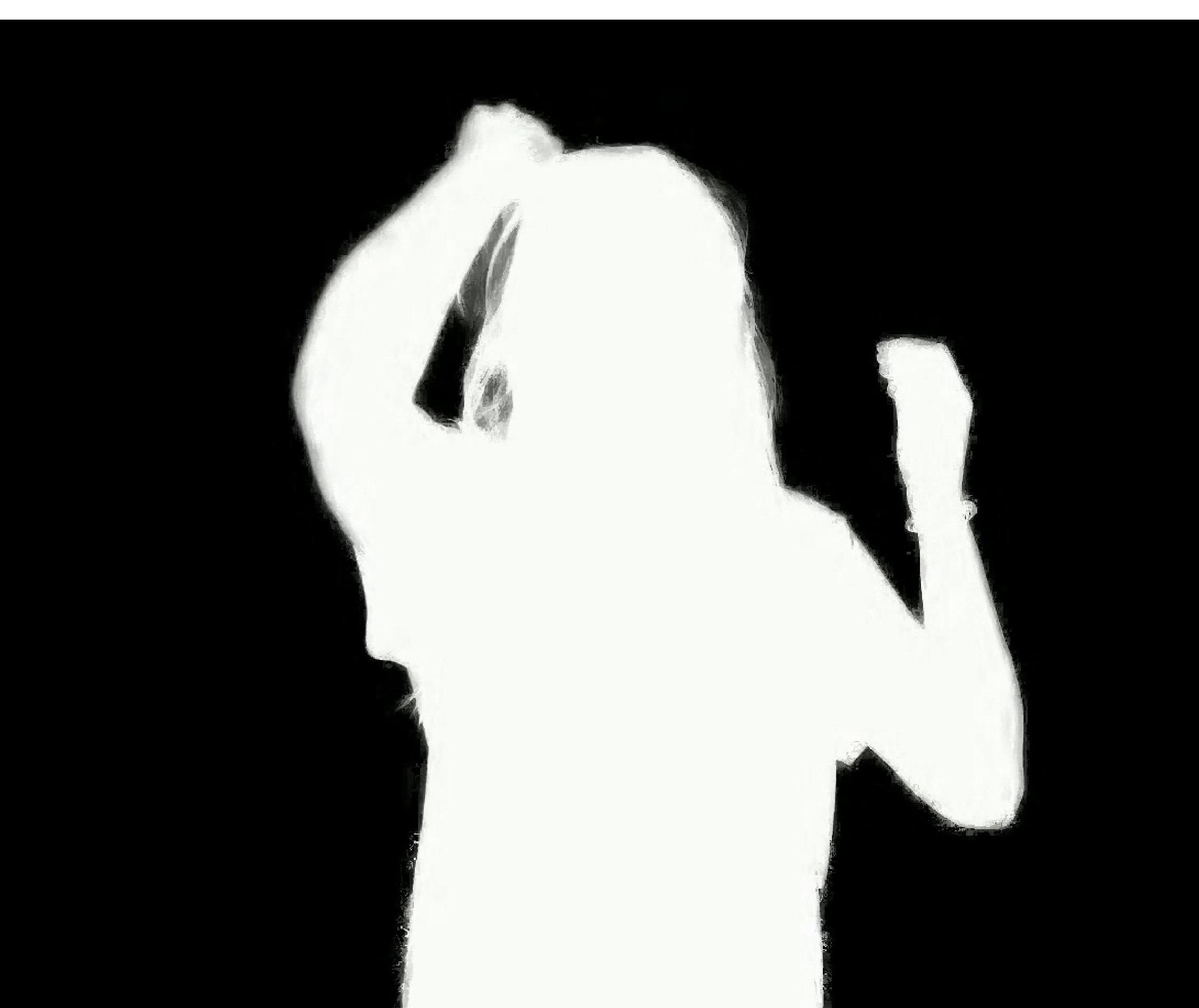} \\
		(c) BGU-fast $8\times$ ($\sim$20fps) & (d) GLU$^- 8\times$ (>30fps) \\
	\end{tabular}
	\vspace{-3mm}
	\caption{Real-time video matting with BackgroundMattingV2~\cite{lin2021real}, which can achieve only less than 5fps with GTX1650 GPU. With the acceleration of our method, it can easily achieve real-time speed.}
	% (c)-(f) are visual comparisons of different upsampling methods. The output boundaries of (c) BGU and (d) BGU-fast are clear, with color bleeding inside the mask. (d) GLU is the most faithful method for small-target, and (f) GLU$^-$ is slightly inferior to glu in preserving details, such as hair ends.}
\label{fig:BVM} 
\end{figure}

\subsection{Interactive Editing with Instant Feedback}
\label{sec:interactive}

As mentioned above, the joint optimization process is target-free, so it can be precomputed for interactive image editing. The optimized upsampling parameter $\Theta$ can be stored as a $3\times H\times W$ tensor. Given $\Theta$, the upsampling requires only a simple linear interpolation for each pixel, whose computation is negligible for interactive applications.

As a demonstration, we built an interactive matting system that enables instant feedback without GPU acceleration. It is known that automatic image matting is very hard for general objects, while interactive matting requires fast speed for better user experience. To demonstrate the power of the proposed approach, we adopted the global matting method proposed in~\cite{levin2007closed}, which needs to solve a large sparse linear system, and is therefore very slow for large images.

Regardless of the size of the input image, our system resizes it to a small image of no more than 10K pixels, for which the matting method is invoked and then the result matte is upsampled to the full resolution for display. Thus, the main computation is to solve the $10000\times 10000$ sparse linear system. We adopted the MKL PARDISO solver, which is very efficient and requires about 0.5 seconds for each solution. Surprisingly, although 10K pixels seems very small, we find it works well with the proposed guided upsampling method. Actually, as shown in \reffig{fig:nice}, our method  can achieve good results for very large ratios such as $64\times$ and $128\times$. For a typical high-resolution input of 10M pixels, downsampling it to 10K requires a downscale ratio about $30\times$, which is not too large in comparison. After a period of user interaction, a trimap can be estimated from the current alpha matte, then our system proceeds with unknown pixels at a finer scale. In this way, the matting operator is invoked from coarse to fine, which helps avoid upsampling errors due to large ratio of downsampling. 

\reffig{fig:interaction} demonstrates the effectiveness of our system. The user can instantly observe changes in the matting result when moving the mouse for editing. This enables the user to add brushes accordingly in the areas with large errors, thus greatly improving efficiency and bringing a much better user experience. Please see the supplementary material for video demonstrations.

%Note that JBU also can be considered as \emph{target-free}, because its weighting function is independent of the target image. However, in order to precompute and store the weights, JBU requires much more memory (e.g. a $25\times H\times W$ tensor for $5\times5$ support), and the upsampling with stored weights is also significantly slower than GLU, so the gain would be much smaller and it should be better to compute the weights on-the-fly.

%Our proposal allows doodle-style user input, and yields satisfactory results based on targets with few input; moreover, we allow users to adjust editing results in real-time to balance the output. As shown in \reffig{fig:simu}, we demonstrate user-interactive doodle targets. Taking alpha matting as an example, we segment the input to simulate the user's doodle actions for matting operator. Our method performs well at edges even with large upsampling scales. Therefore, proposed GLU can reconstruct the complete alpha mask well even if we treat a few simple user doodles as the target image. 

\subsection{Real-time Video Processing}
\label{sec:realtime}

Our method is parallelable and can be executed fast with GPU implementation, thus enabling real-time video processing. To demonstrate this, we apply our method to BackgroundMattingV2~\cite{lin2021real}, an efficient video matting method that can achieve 30fps for 4K videos on RTX2080ti GPU. However, for our low-end test machine with GTX1650 GPU, it requires more than 200ms for each frame. To achieve real-time speed, we apply our method with 8$\times$ downsampling, then the low-resolution matting with BackgroundMattingV2 requires about 21ms, and upsampling with GLU$^-$ requires about 14ms, so in total it takes about 35ms per frame.

The target-free property of our method can further improve the efficiency. Since the optimization of GLU$^-$ is independent of the  matting result, it can be executed in parallel with the matting procedure. In this way the time cost can be further reduced. Compared to the results in full resolution, the sacrifice in quality is usually small for 8$\times$ downsampling, as shown in \reffig{fig:BVM}. In comparison with BGU-fast, our method can achieve faster speed and better quality.

\begin{figure}[ht]
	\centering 
	\subfigure[Source]{
		\includegraphics[width=0.31\linewidth]{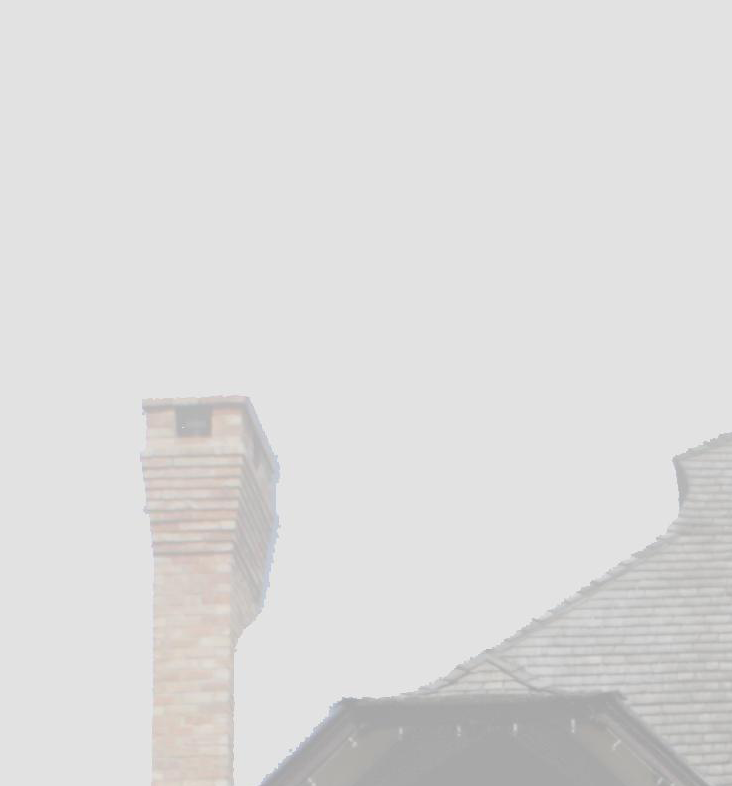}
	} \hspace{-1.5mm}
	\subfigure[Reference]{
		\includegraphics[width=0.31\linewidth]{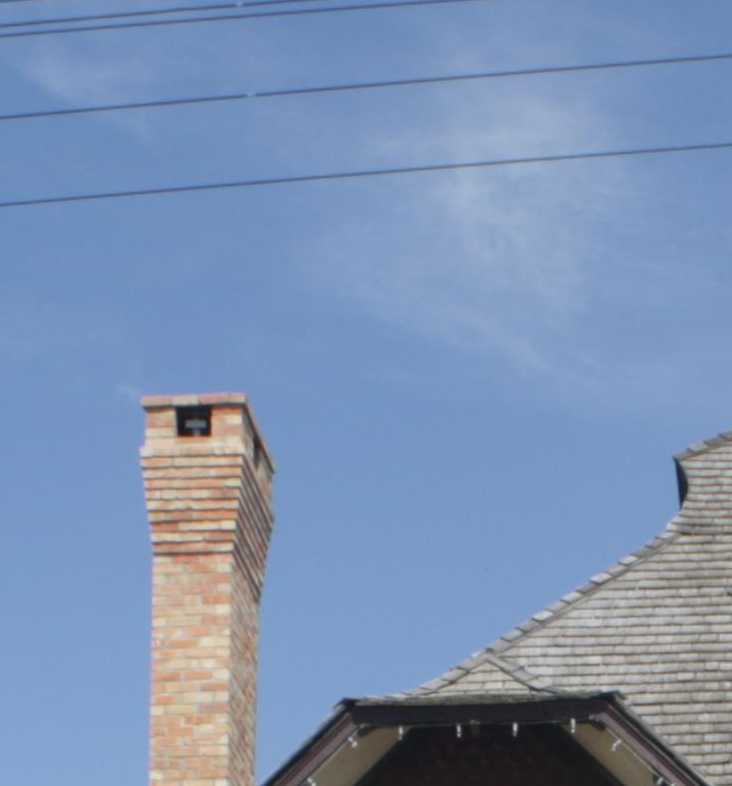}
	} \hspace{-1.5mm}
	%	\subfigure[$16\times\uparrow$BGU]{
		%		\includegraphics[width=0.23\linewidth]{figure/limit1/bgu.png}
		%	} \hspace{-1.5mm}
	\subfigure[GLU $8\times$ ]{
		\includegraphics[width=0.31\linewidth]{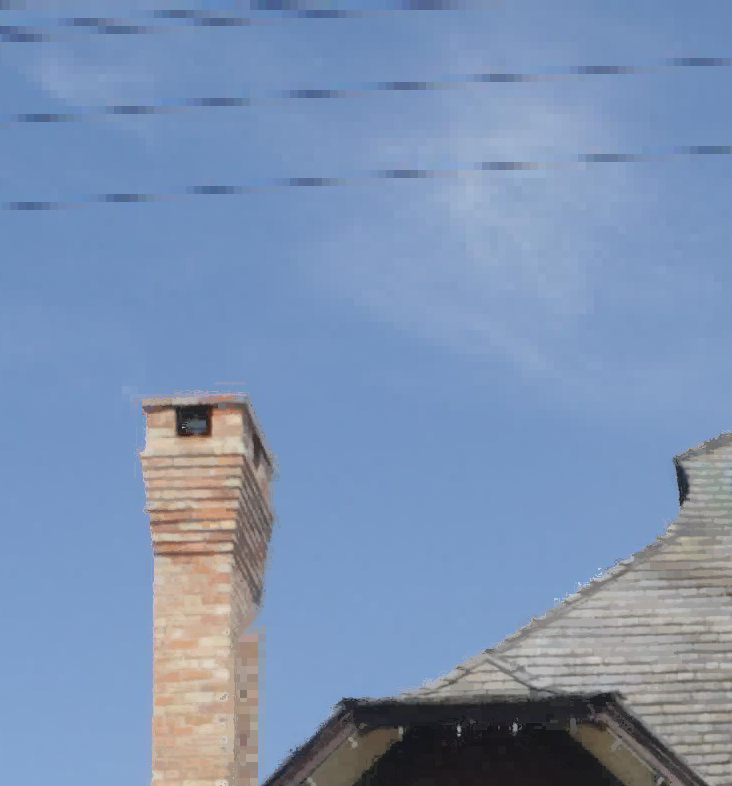} 
	} 
	\vspace{-3mm}
	\caption{Failure case of our method. New edges not appear in the source image cannot be well recovered.}
	\label{fig:limit1} 
\end{figure}

\begin{figure}[ht]  
	\centering 
	\subfigure[Source]{
		\includegraphics[width=0.31\linewidth]{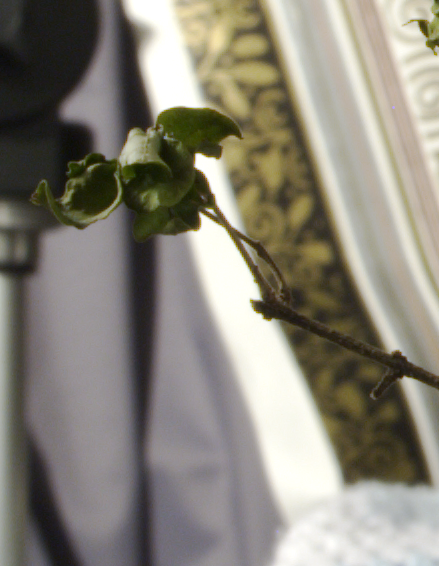}
	} \hspace{-1.5mm}
	\subfigure[Reference]{
		\includegraphics[width=0.31\linewidth]{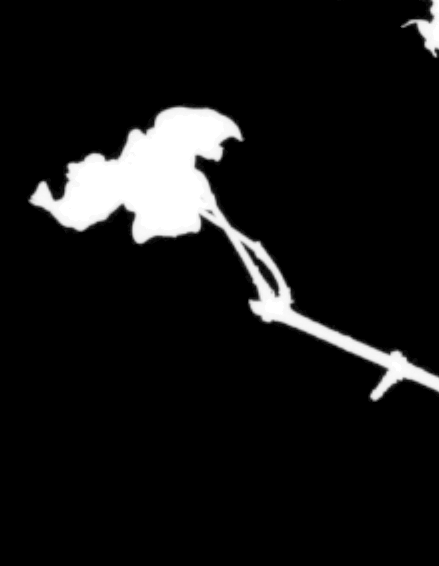}
	} \hspace{-1.5mm}
	%	\subfigure[$16\times\uparrow$BGU]{
		%		\includegraphics[width=0.23\linewidth]{figure/limit2/bgu8.png}
		%	} \hspace{-1.5mm}
	\subfigure[GLU $8\times$]{
		\includegraphics[width=0.31\linewidth]{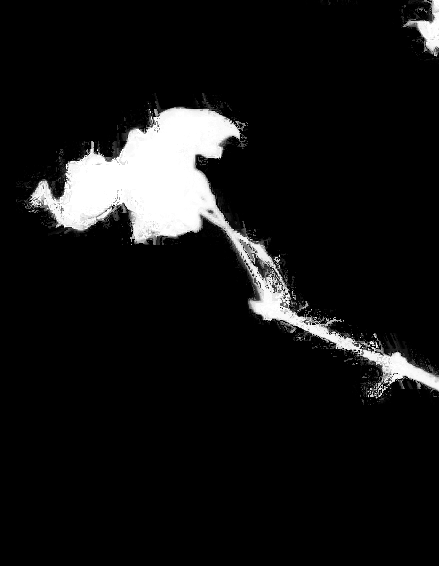} 
	} 
	\vspace{-3mm}
	\caption{Our method may produce unsmooth artifacts when the source and target images have very different pixel affinities.}
	\label{fig:limit2} 
\end{figure}

\begin{figure*}[t]   
	\centering
	\includegraphics[width=\linewidth]{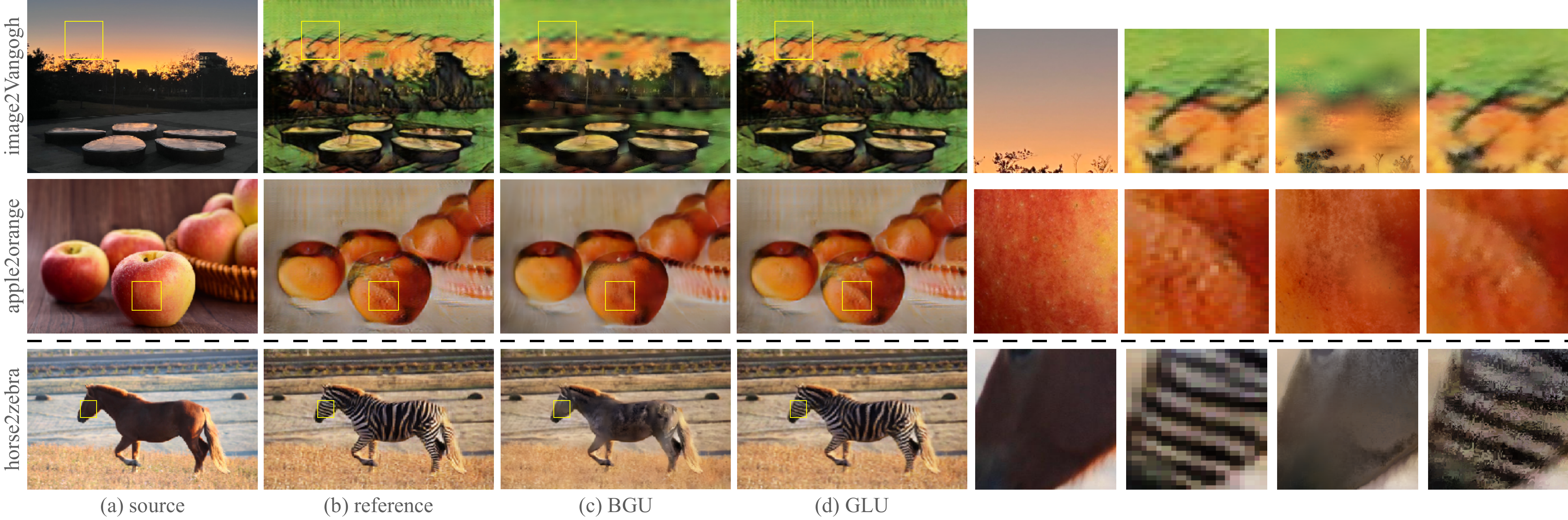}
	\caption{Examples of CycleGAN style transfer~\cite{CycleGAN2017}, which may introduce dramatic changes to local image structures. In this case, our method may produce unsmooth artifacts, while BGU may smooth out the new structures. Since CycleGAN can only output  256$\times$256 fixed-size results, the reference here is the low-resolution target image.}
	\label{fig:learning-compare}
\end{figure*}

\section{Limitations}
\label{sec:limitations}

The same as JBU and BGU, a basic assumption of our method is that the source and target images have almost the same local affinity, i.e. pixels with more similar color in the source image should have more similar color in the target output. Therefore, our method is not suitable for applications that may introduce \emph{new edges} in the target image.  \reffig{fig:limit1} shows such an example. The affinity of new edges differs a lot from the source and is therefore out of the scope of joint optimization. In a more general case, our method may produce unsmooth edges due to different local affinities. As shown in \reffig{fig:limit2}, since the object has similar color to the background, our method fails to accurately recover the sharp matte edges. 

%\revised{Another interesting application might be \textbf{Optical Flow}, we test with the RAFT flow method \cite{teed2020raft} and test images from the MPI Sintel dataset~\cite{Butler:ECCV:2012}, whose image size is small ($1024\times 468$), so we test with only smaller downsampling ratio of $4\times$. As shown in \reffig{fig:results}, our method produces unsmooth edges.}

Note that unlike adding new edges, removing edges does not usually cause problems for our method, because the resultant pixels can still be interpolated by neighboring pixels, so applications such as matting and smoothing can be well supported. In contrast, BGU may insistent on preserving the edges and local structures of the source image, so for edge-removing applications it may produce artifacts, as demonstrated in \reffig{fig:results}.

Due to the limitation on handling new edges, our method is not suitable for applications that may drastically change the local image structures, such as the recent learning-based style transfer methods~\cite{CycleGAN2017,park2020swapping}. This is a common limitation for universal guided upsampling methods including JBU and BGU. \reffig{fig:learning-compare} shows some examples. As shown, for cases such as \emph{Horse2Zebra}, BGU may completely ignore new edges, while our method results in unsmooth artifacts. In fact, since for new edges, there is no guidance information in the source image, it is not able to get good results without any domain-specific prior knowledge, so using learning-based approaches should be better in this case.

%be better to be upsampled with normal resampling methods such as bilinear or bicubic interpolations. Therefore, an immediate way to improve our method for such applications is to detect new edges and then adaptively incorporate GLU with bicubic interpolation, which is not difficult to achieve with some more computations for the interpolation process.
For real applications, a practical issue is to find an image operator that is suitable for low-resolution processing. As demonstrated in \reffig{fig:app-um}, the input image scale may have a great effect on the output of some image processing methods. For example, we found that BackgroundMattingV2 produces significantly more errors when being applied to $8\times$ downsampling inputs, because it is originally designed for high-resolution video. 
%\revised{Another example is optical flow estimation~\cite{teed2020raft}, whose accuracy is greatly influenced by the image resolution, and may need a special refinement process after the upsampling.} 
\onerevise{Another example is optical flow estimation~\cite{teed2020raft}, whose accuracy is greatly influenced by image resolution and requires special treatment after upsampling.}
How to adapt these methods for better low-resolution image processing is a remaining problem.

\section{Conclusion}

We propose a simple yet effective guided upsampling method, which represents each high-resolution pixel as a linear interpolation of  two low-resolution pixels. Transition edges and smooth variations in natural images can be well represented. The upsampling parameters and the downscaled input image are jointly optimized in order to minimize the upsampling error. We reveal and discuss the connections with previous methods. In particular, our method can be considered as a special case of JBU with optimized weights, and it also implicitly exploits pixel-level local color transformations. These properties enable it to overcome the blurring and bleeding artifacts of previous approaches.

We demonstrate the advantages of our method with a wide range of image operators. For interactive editing tasks, our approach enables time-costly operators to achieve instant feedback without hardware acceleration. Real-time video processing can also be enabled easily for high-resolution inputs. Finally, our work shows that a high-resolution image can be well represented by simple linear interpolation of its downscaled counterpart. Similar to bilateral filter and local color transformations, the power of such representation can be further explored in related tasks such as learning-based image processing~\cite{gharbi2017deep}, super resolution~\cite{sun2020learned}, segmentation~\cite{mazzini2018guided}, etc.

\section{Acknowledge}
The authors would like to thank the anonymous reviewers for their valuable comments and suggestions. This work is supported by National Key R\&D Program of China under grant (2022YFB3303200), Natural Science Foundation of China (62072284), and Center-initiated Research Project of Zhejiang Lab (2021NB0AL01).

\bibliographystyle{ACM-Reference-Format}
\bibliography{reference} 

\end{document}